%% file: main.tex
\documentclass[akbc,twoside,11pt,lettersize]{article}

\usepackage{akbc}
\akbcheading
\usepackage[utf8]{inputenc} 
\usepackage[T1]{fontenc}    
\usepackage{hyperref}       
\usepackage{url}            
\usepackage{booktabs}       
\usepackage{amsfonts}       
\usepackage{nicefrac}       
\usepackage{microtype}      

\usepackage{graphicx}
\usepackage{epsfig}
\usepackage{epstopdf}
\usepackage{adjustbox}
\usepackage{subfig}

\usepackage{comment}
\usepackage{microtype}
\usepackage{multirow}
\usepackage{graphicx}
\usepackage{float}
\usepackage{adjustbox}
\usepackage{physics}

\usepackage{xcolor}
\usepackage{dblfloatfix}
\usepackage{soul}
\usepackage{booktabs}
\usepackage{relsize}
\usepackage{xcolor,colortbl, makecell}
\usepackage{enumitem}
\usepackage[normalem]{ulem}
\useunder{\uline}{\ul}{}
\usepackage{tcolorbox}
\usepackage{diagbox}

\usepackage{amssymb}
\usepackage{pifont}

\newcommand{\cmark}{\ding{51}}%
\newcommand{\xmark}{\ding{55}}%

\newcommand\jeff[1]{{\color{blue}\{\textit{#1}\}$_{jeff}$}}
\newcommand\ronan[1]{{\color{teal}\{\textit{#1}\}$_{ronan}$}}
\newcommand\antoine[1]{{\color{orange}\{\textit{#1}\}$_{antoine}$}}

\newcommand{\atomicTT}{\textsc{Atomic$^{20}_{20}$}}
\newcommand{\COMETTTO}{\textsc{COMeT}}
\newcommand{\COMET}{\textsc{COMeT}}
\newcommand{\eg}{\textit{e.g.}}
\newcommand{\ie}{\textit{i.e.}}

\title{Analyzing Commonsense Emergence \\ in Few-shot Knowledge Models}

\akbcfinaltrue

\newcommand{\ai}{$^1$}
\newcommand{\epfl}{$^2$}
\newcommand{\uw}{$^{3}$}
\newcommand{\aiepfl}{$^{1,2}$}
\newcommand{\aiuw}{$^{1,3}$}

\author{\name Jeff Da\ai \email jeffd@allenai.org \\ 
        \name Ronan Le Bras\ai \email ronanlb@allenai.org \\ 
        \name Ximing Lu\ai \email ximinglu@allenai.org \\
        \name Yejin Choi\aiuw \email yejinc@allenai.org \\ 
        \name Antoine Bosselut\aiepfl \email antoine.bosselut@epfl.ch \\ 
        \addr \ai{}Allen Institute for AI \\
        \addr \epfl{}EPFL \\
        \addr \uw{}University of Washington
        }

\begin{document}

\maketitle

\input{sections/0-abstract}
\input{sections/1-introduction}

\input{sections/2-background}

\input{sections/3-experimental}

\input{sections/4-model-studies}
\input{sections/5-analysis}
\input{sections/6-related}
\input{sections/7-conclusion}

\section*{Acknowledgements}
{
The authors would like to thank the anonymous reviewers for their feedback, and the Amazon Mechanical Turk community for help with annotation. The authors thank 
Vered Shwartz and Chris Manning for feedback on early drafts. We also gratefully acknowledge the support of DARPA under No. N660011924033 (MCS), JD.com, and the Allen Institute for AI. TPU machines for conducting experiments were provided by Google.
}

\bibliography{emnlp2020}
\bibliographystyle{plainnat}

\newpage

\input{sections/8-appendix}

\end{document}

%% file: sections/0-abstract.tex


\begin{abstract}
Recently, commonsense knowledge models --- pretrained language models (LM) fine-tuned on knowledge graph (KG) tuples --- showed that considerable amounts of commonsense knowledge can be encoded in the parameters of large language models \cite{Bosselut2019COMETCT}.
However, as parallel studies show that LMs are poor hypothesizers of declarative commonsense relationships \cite{Petroni2019LanguageMA} on their own, it remains unclear whether this knowledge is learned during pretraining or from fine-tuning on KG examples.

To investigate this question, we train commonsense knowledge models in few-shot settings to study the emergence of their commonsense representation abilities. Our results show that commonsense knowledge models can rapidly adapt from limited examples, indicating that KG fine-tuning serves to learn an interface to encoded knowledge learned during pretraining. Importantly, our analysis of absolute, angular, and distributional parameter changes during few-shot fine-tuning provides novel insights into \textit{how} this interface is learned. 




\end{abstract}

%% file: sections/1-introduction.tex
\section{Introduction}
\label{sec:intro}



Driven by advances in pretrained language models \cite{Liu2019RoBERTaAR,Raffel2019ExploringTL}, recent NLP systems have demonstrated considerable improvement \cite{Clark2020TransformersAS,Lourie2021UNICORNOR} on benchmarks that test the commonsense representation and reasoning abilities of NLP systems \cite{talmor-etal-2019-commonsenseqa,sap-etal-2019-social,Bhagavatula2020AbductiveCR,Sakaguchi2020WINOGRANDEAA}. Despite this success, it remains unclear how much commonsense knowledge is actually directly encoded by these models. When prompted out-of-the-box to complete commonsense declarative relationships (as depicted in Figure~\ref{fig:intro}), they exhibit limited ability to map their language modeling abilities to this task \cite{Petroni2019LanguageMA,Zhou_Zhang_Cui_Huang_2020}.

However, the zero-shot expression of declarative commonsense knowledge is but one use of language that these models are pretrained to manifest. Infilling evaluations that entangle this ability with general language expression may be narrow tests of true commonsense representation ability. Indeed, commonsense knowledge models \cite{Bosselut2019COMETCT} --- which finetune pretrained LMs on examples from commonsense knowledge graphs \cite{Speer2017ConceptNet5A,Sap2019ATOMICAA,Jiang2021ImNM} --- learn to express declarative commonsense relationships much more effectively.
This learning procedure allows these models to recalibrate to the actual task of hypothesizing declarative knowledge. Simultaneously, these systems are evaluated in challenging train/test settings (where head entities tested on cannot be seen during training), indicating they may learn to transfer implicit representations of declarative knowledge that are learned during pretraining \cite{Hwang2020COMETATOMIC2O}.

\input{figures/fig1}

However, current commonsense knowledge models are trained on a large number of examples from knowledge graphs. Consequently, this comprehensive finetuning obfuscates whether these systems exhibit knowledge that was encoded during language pretraining \cite{Geva2020TransformerFL,Dai2021KnowledgeNI}, or whether pretraining merely provides a favorable initialization to learn to generalize from examples in the KG. In this work, we explore this question by training commonsense knowledge models in a few-shot setting. We vary the training budgets of KG examples that they receive, and evaluate their capacity to hypothesize commonsense knowledge after training on each of these budgets. Using this framework, we discover whether the finetuning process causes the model to learn new knowledge from the KG, or whether the model learns an \textit{interface} to existing knowledge it already encodes.


Our results demonstrate that few-shot commonsense knowledge models trained on knowledge graphs in few-shots --- up to 10,000$\times$ fewer tuples than the full KG --- exhibit strong performance relative to zero-shot LMs, and approach the performance of fully-trained knowledge models. Our analysis supports the hypothesis that few-shot finetuning allows the model to adapt knowledge it already encodes implicitly. 
We find that few-shot training mainly changes parameters in the attention heads of the decoder transformer blocks. The large feed-forward networks that serve as an implicit memory storage for information learned during pretraining \cite{Geva2020TransformerFL} are minimally updated. Furthermore, we observe that larger knowledge models exhibit most of their parameter change across a more concentrated set of parameters, implying they learn knowledge during pretraining in a less entangled manner due to their capacity. Finally, we find that using natural language prompts to represent relation inputs accelerates commonsense emergence.
Our code, few-shot training sets, and models are at \href{https://github.com/allenai/few-shot-comet/}{https://github.com/allenai/few-shot-comet/}. 

%% file: figures/fig1.tex
\begin{figure}
\centering
  \includegraphics[width=\textwidth]{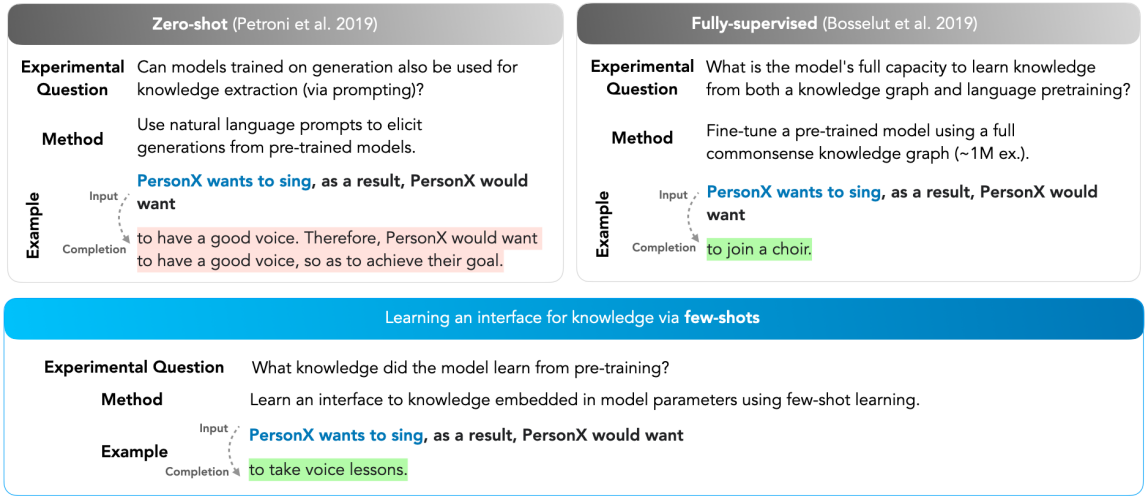}
  \caption{Commonsense knowledge models can be trained to effectively hypothesize commonsense knowledge when trained in few-shots, implying that finetuning serves to \textit{learn} and interface to knowledge encoded during pretraining. 
  }
  \label{fig:intro}
\end{figure}

%% file: sections/2-background.tex
\section{Background}

In this work, we investigate few-shot performance and parameter change of commonsense knowledge models to shed light on the dynamics of commonsense representation learning. Below, we describe background concepts to contextualize this analysis.

\noindent \textbf{Commonsense Knowledge Graphs} are structured, relational representations of commonsense knowledge. 
In our study, we use \atomicTT{} \cite{Hwang2020COMETATOMIC2O},\footnote{Downloadable at \url{https://allenai.org/data/atomic-2020}} a commonsense KG with 1.33M inferential knowledge tuples about entities and events. It represents a large-scale commonsense repository of textual descriptions that encode social and physical aspects of common human experiences. Across its 1.33M tuples, it captures information about 23 relationship types.
Example head entities and relations are shown in Table~\ref{table:generations}. \atomicTT{} is split into training, development, and test subsets such that no head entities in one set appear in any other. This property allows models trained on this resource to be evaluated on their capacity to generalize commonsense relationships to new entities, events, and situations.

\input{figures/generations}

\noindent \textbf{Commonsense Knowledge Models} represent facts by learning to encode a commonsense KG~\cite{Bosselut2019COMETCT}. 
They are pre-trained as language models and fine-tuned on knowledge graph tuples to learn to hypothesize knowledge relationships through language generation. After training on a large collection of tuples from a KG, they learn the structure and relationships encoded by that KG. Furthermore, because they are seeded with pretrained language models, they learn to generalize the relationships to other entities about which the LM implicitly encodes knowledge \cite{Petroni2019LanguageMA,roberts-etal-2020-much}. Consequently, they can be used to produce precise knowledge on-demand for any entity that can be expressed through language. 

\input{figures/example_prompts}

%% file: figures/generations.tex
\begin{table*}[t]
\resizebox{\linewidth}{!}{
\begin{tabular}{llll}
\toprule
\textbf{Head $h$} & \textbf{Relation $r$} & \textbf{\makecell[l]{Generated Tail $t$ (\COMETTTO)}} & \textbf{\makecell[l]{Generated Tail $t$ (GPT-3)}} \\ \toprule
nail & \textsc{AtLocation} & in wall \cmark & hammer \xmark \\ \hline
video camera & \textsc{ObjectUse} & video recording \cmark & record the grooming session \cmark \\ \hline
PersonX takes it to the vet & \textsc{HinderedBy} & \makecell[l]{PersonX doesn't have time \cmark} & \makecell[l]{PersonX doesn't know where the vet is. \cmark} \\ \hline
\makecell[l]{PersonX gets a call for an interview} & \textsc{xAttr} & \makecell[l]{PersonX is qualified for the job \cmark} & lucky \cmark \\
\makecell[l]{PersonX wants to learn how to swim} & \textsc{xAttr} & \makecell[l]{PersonX isn't confident in the water \cmark} & a good idea \xmark \\ \hline
PersonX falls ill & \textsc{xEffect} & PersonX will miss work \cmark & \makecell[l]{is in bed \cmark} \\
PersonX sprays by a skunk & \textsc{xEffect} & PersonX will be sick \cmark & is stinky \cmark \\ \hline
PersonX misses class & \textsc{xNeed} & to have a valid excuse \cmark & to have a class \cmark \\ \hline
PersonX notices a strange smell & \textsc{xWant} & to investigate the smell \cmark & \makecell[l]{PersonX is distracted by the\\smell of PersonY's perfume. \xmark} \\
PersonX wants to learn how to sing & \textsc{xWant} & to take voice lessons \cmark & to learn how to play an instrument \xmark \\ \bottomrule
\end{tabular}
}
\caption{Examples of few-shot ($n=3$) generations produced by \COMETTTO{} (T5) and GPT-3. \COMETTTO{} (T5) produces diverse and novel tail hypotheses despite learning from few examples and is able to learn the task interface more easily than GPT-3, which tends to copy the input or generate unnecessary full sentences. Assessments (\cmark, \xmark) are from human evaluators. 
}
\label{table:generations}

\end{table*}

%% file: figures/example_prompts.tex
\begin{figure}[!tbp]
  \centering
  \begin{minipage}[b]{0.48\textwidth}
    \begin{adjustbox}{width=\textwidth}
    \centering
    \begin{tabular}{ll}
    \textbf{Relation $r$} & \textbf{Prompt} \\ \toprule
    \textsc{ObjectUse} & $h$ is used for $t$ \\
    \textsc{AtLocation} & You are likely to find a $h$ in a $t$ \\
    \textsc{xIntent} & Because of $h$, PersonX wanted $t$ \\
    \textsc{xWant} & After $h$, PersonX would want $t$. \\
    \textsc{xAttr} & $h$ is seen as $t$ \\
    \textsc{isAfter} & Something that happens after $h$ is $t$ \\
    \textsc{oWant} & As a result of $h$, others would want $t$ \\ \bottomrule
    \end{tabular}
    \end{adjustbox}
    \captionof{table}{Examples of language prompts used to represent relations for knowledge models. Prompts significantly speed up transfer in few-shot learning settings.}
    \label{table:example_templates}
  \end{minipage}
  \hfill
  \begin{minipage}[b]{0.48\textwidth}
  \centering
  \includegraphics[width=\textwidth]{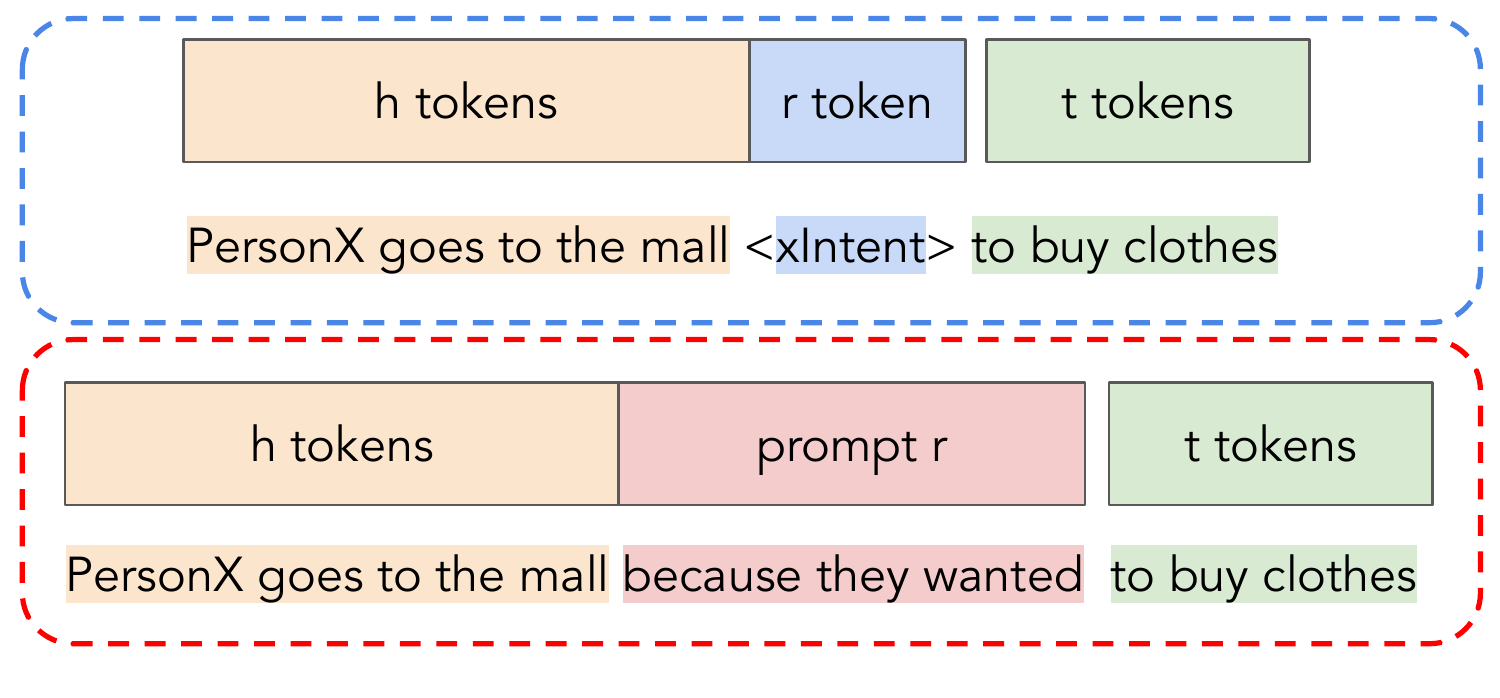}
  \caption{With prompting, KG relations are mapped to natural language templates that more closely resemble inputs the model has learned to process from pretraining.}
  \label{fig:example_templates}
  \end{minipage}
\end{figure}

%% file: sections/3-experimental.tex
\section{Experimental Setup}
\label{sec:method}

In this section, we outline our experimental setting for training few-shot knowledge models.

\paragraph{Input}
 Each tuple in the \atomicTT{} knowledge graph is composed of a {\{head \textit{h}, relation \textit{r}, tail \textit{t}\}} triplet. The head and tail entities in the triplet are natural language words or phrases. Commonsense knowledge models are trained by providing the tokens of \textit{h} and \textit{r} as inputs to the model and learning to generate the tokens of \textit{t}. In this work, rather than initializing \textit{r} with a new token and random embedding \cite{Bosselut2019COMETCT}, we automatically format the input tuples into natural language prompts to represent the relation \cite{Feldman2019CommonsenseKM,jiang20tacl}. Table \ref{table:example_templates} shows examples of such prompts.



\vspace{-1ex}
\paragraph{Training}

We source training tuples from the \atomicTT{} training set \cite{Hwang2020COMETATOMIC2O}. When constructing few-shot training sets, we set a target number $n$ of examples to randomly sample from the knowledge graph for each relation (\ie, $n = 3 \implies 3$ examples $\times~23$ relations = $69$ total training examples). Once a training set of examples is produced, the model is trained on this subset to minimize the negative log-likelihood of the tokens of the tail entity for each tuple. We use the AdaFactor optimizer \cite{Shazeer2018AdafactorAL} with a constant learning rate of 0.001, a mini-batch size of 4, and train the model for 3 epochs. 
Unless stated otherwise (\S\ref{ssec:modelsize}), we use T5-11B \cite{Raffel2019ExploringTL} as a seed language model for all experiments. 
To pick hyperparameters during fine-tuning, we run 5 different sets of hyperparameters and use a validation set of equivalent size to the training set ($n$ examples / relation) to evaluate performance \cite{Perez2021TrueFL}.

\vspace{-1ex}
\paragraph{Evaluation}

We evaluate the knowledge hypothesized by the trained few-shot models using human and automatic evaluations.  For the human evaluation (Accept \% in Table \ref{table:firsttable}), we use the procedure described in \cite{Hwang2020COMETATOMIC2O}. We ask annotators to label the plausibility generated tuples using a 4-point Likert scale: \{\textit{always/often true $(+2)$, sometimes/likely true $(+1)$, false/untrue $(-1)$, nonsensical $(-2)$}\}. 
We collect 3 annotations per relation, convert each annotation to an acceptability label (\ie, $\{+1,+2\} \rightarrow$ \cmark, $\{-1,-2\} \rightarrow$ \xmark) and use the majority label as the acceptability judgment. Evaluation agreement is measured using Fleiss's $\kappa = 0.49$ 
for the acceptability judgment. 
We also evaluate hypothesized knowledge tuples using automatic metrics: BLEU-1  \cite{Papineni2002BleuAM}, METEOR \cite{Banerjee2005METEORAA}, ROUGE-L \cite{Lin2004ROUGEAP}, and CIDEr \cite{Vedantam2015CIDErCI}. For few-shot experiments, we report average performance across 5 runs with different training sets.


\input{figures/firsttable}

%% file: figures/firsttable.tex
\begin{table*}[t]
\begin{adjustbox}{width=\textwidth}
\begin{tabular}{llrrrrrr}
\textbf{Methodology} & \textbf{Model} & \textbf{BLEU-1} & \textbf{METEOR} & \textbf{ROUGE-L} & \textbf{CIDEr} & \textbf{Accept \%} \\ \toprule
\multirow{2}{*}{zero-shot} & GPT-2 XL & 10.1 & 8.2 & 9.8 & 4.7 & 36.6 \\
& GPT-3 & 14.3 & 11.7 & 13.9 & 4.7 & 39.9 \\ \midrule
\multirow{4}{*}{\makecell{few-shot\\($n = 3$)}} & GPT-2 XL (augmentation) & 15.1 & 10.2 & 13.5 & 6.5 & 37.5 \\
 & \COMETTTO~(GPT-2 XL) (finetuning) & 18.4 & 11.1 & 14.2 & 7.3 & 40.1 \\
 & GPT-3 (augmentation) & 29.9 & 17.8 & 25.3 & 19.0 & 71.7 \\
 & \COMETTTO~(T5) (finetuning) & 24.2 & 14.4 & 21.3 & 18.1 & 75.7 \\ \midrule
\multirow{2}{*}{fully} & \COMET{} (GPT-2 XL) & 40.7 & 29.2 & 48.5 & 65.3 & 72.5 \\ \multirow{1}{*}{\rule{0pt}{3ex}supervised}
 & \COMET{} (BART) & 46.9 & 33.0 & 49.5 & 65.8 & 84.5 \\
 & \COMETTTO{} (T5) & \bf{48.2} & \bf{34.1} & \bf{50.0} & \bf{66.4} & \bf{86.4} \\ \bottomrule
\end{tabular}
\end{adjustbox}
\caption{Comparison between various trained knowledge models. Few-shot ($n = 3$) knowledge models transfer well in both the finetuning (\COMET{}) and augmentation (GPT-3) settings.}
\label{table:firsttable}
\end{table*}

%% file: sections/4-model-studies.tex
\vspace{-1.5ex}
\section{Do few-shot knowledge models learn?}
\label{sec:modeling_studies}

We evaluate the general few-shot commonsense interface learning capability of large-scale language models. 
%
%
%
%
%
We train \COMETTTO{} models as described in \S\ref{sec:method} and set the number of training examples per relation to $n = 3$. We report several zero-shot and few-shot augmentation baselines (\eg, GPT-\{2 XL, 3\}), which prepend $n$ training examples (for the same relation) to the start of the input sequence as a prompt, and condition the generation on these additional examples. Few-shot augmentation baselines are not fine-tuned, but receive examples using the same prompt formats as the finetuning baselines (Tables~\ref{table:example_templates},~\ref{table:example_templates_large}). Similar to fine-tuning, we report average performance across 5 different example sets used for prompting. 
As T5 models are not pretrained using a causal language modeling objective, reporting zero-shot evaluation performance for a multi-word generation task would be an unfair comparison, so we only report limited numbers in Table~\ref{tab:t5-zero} of Appendix~\ref{sec:app:t5}. 
We also report the results of several fully supervised \COMETTTO{} knowledge models.
\paragraph{Findings}
We find that both the few-shot augmentation and few-shot learning settings are able to produce high quality commonsense knowledge tuples, indicating that large LMs are also efficient few-shot learners (in addition to showcasing impressive few-shot prompting abilities). Using only $n = 3$ tuples per relation, both GPT-3 and \COMETTTO~(T5) produce high-quality tuples that are accepted as \textit{plausible} by human evaluators more than 70\% of the time --- 75.7\% for \COMETTTO~(T5). While this performance falls short of the fully supervised knowledge model performance --- 86.4\% --- it  demonstrates that the commonsense representation abilities of language models extends much further than their zero-shot prompting abilities. Furthermore, these abilities can be measured by finetuning on limited examples, supporting the hypothesis that early-stage finetuning serves to learn an interface to knowledge learned during pretraining. In the following section, we support this hypothesis by investigating how parameters of large language models change during finetuning (across different training budgets $n$ \S\ref{ssec:paramchange}, different model size \S\ref{ssec:modelsize}, and different prompting strategies \S\ref{ssec:prompts}).

%% file: sections/5-analysis.tex
\section{How do knowledge models learn?}

Despite strong empirical results showing that pretrained language models can rapidly adapt to become few-shot language learners \cite{Schick2020ItsNJ}, little work has focused on \textit{how} language models learn to adapt knowledge learned from pretraining to perform well on downstream end tasks. We explore this problem in the context of few-shot knowledge modeling and investigate how the parameters of few-shot commonsense knowledge models change during finetuning. 
We define three measures to investigate parameter change: absolute parameter change, angular parameter change, and distribution of parameter change.

\subsection{Measuring parameter change}
\label{ssec:param_change}

\paragraph{Notation}

To more easily discuss measures of parameter change during knowledge model fine-tuning, we review the structure of the transformer blocks in an encoder-decoder transformer LM \cite{Vaswani2017AttentionIA}. Encoder blocks consist of a self-attention layer followed by a feedforward network. Each self-attention layer contains parameter matrices (referred to in the following sections as $q$, $k$, $v$) that project the query, key, and value inputs before computing a scaled dot-product attention between the key and value, and an output projection $o$ of the attended value vectors. The trainable parameters in the feedforward network consist of two linear projection matrices $wi$, $wo$. Each decoder block consists of the same parameter matrices with an additional cross-attention layer (parameter matrices $xq$, key $xk$, value $xv$, and output $xo$) that attends to the outputs of the encoder blocks. 


\paragraph{Absolute Parameter Change} To measure a normalized quantity for average change in each parameter matrix, we compute a normalized $\ell_1$ distance for each set of parameter matrices in the transformer blocks. Without loss of generality, for each of these parameter matrices, we define $\Theta^{PT}$ as the matrix of parameters before finetuning, and $\Theta^{FT}$ as the matrix of parameters post-finetuning.
The normalized $\ell_1$ distance between these two matrices for each parameter matrix type is then: 
\par\nobreak
\begin{small}
\begin{equation}
    d_{\ell_1} = \frac{1}{mn} {\norm{\Theta^{FT} - \Theta^{PT}}_1},
\end{equation}
\end{small}
\par\nobreak
\noindent where $m$ and $n$ correspond to the row and column dimensionality of the parameter matrix, and $\norm{\cdot}_1$ corresponds to the $\ell_1$ norm. 
Intuitively, $d_{\ell_1}$ measures the degree of absolute per parameter shift between these two sets of parameters. 

\paragraph{Angular Parameter Change} However, measuring normalized $\ell_1$ distance solely captures how the magnitude of the parameter matrix changes, but obfuscates relative drift between parameters in each matrix. For instance, a matrix with high average L1 change may in fact have small relative changes between its elements  (i.e., the parameter change may mainly cause a scaling effect on the output vectors). Consequently, we measure the row-wise angular distance between $\Theta^{PT}$ and $\Theta^{FT}$ to capture the rotational drift of parameters during finetuning:
\par\nobreak
\begin{small}
\begin{equation}
    d_{ang} = \frac{1}{m\pi}\sum_{k=1}^{m}\cos^{-1}\Bigg(\frac{\langle\Theta_k^{FT}, \Theta_k^{PT}\rangle}{\norm{ {\Theta_k^{FT}}}_2 \norm{{\Theta_k^{PT}}}_2}\Bigg),
\end{equation}
\end{small}
\par\nobreak
\noindent where $k$ indexes a common row vector in both parameter matrices, $\langle \cdot \rangle$ corresponds to an inner product, $\norm{\cdot}_2$ is the Euclidean ($\ell_2$) norm, and $\cos^{-1}$ is the inverse cosine operation.
Contrary to $d_{\ell_1}$, which represents a per-parameter average distance, $d_{ang}$ is computed between all corresponding rows of the parameter matrices, $\Theta^{PT}$ and $\Theta^{FT}$, and averaged per row. We choose to average rows as opposed to columns as each row in the parameter matrix performs a dot product with the input vector to the transformation, and has a correspondence to an individual element in the output vector.

\input{figures/auc}

\paragraph{Distributional Parameter Change} In additional to angular distance, we also report the distribution of parameter changes for a given matrix. More concretely, parameter changes may be concentrated across varying numbers of dimensions. We compute the change in each dimension by calculating $W = |\Theta^{PT} - \Theta^{FT}|$, sorting parameters by their absolute change, and plotting \% cumulative parameter number against \% cumulative mass as a function of a given parameter change $w_i \in W$:
\par\nobreak
\begin{small}
\begin{equation} \label{eq:auc}
\Bigg(\frac{\texttt{count}(w \in W, w \leq w_i)}{\texttt{count}(w \in W)}, \frac{\sum_{w \in W, w \leq w_i} w}{\sum_{w \in W} w}\Bigg)
\end{equation}
\end{small}
\par\nobreak
\noindent Figure \ref{fig:auc} visualizes this relationship. 
Intuitively, a matrix with a small AUC has a skewed relative distribution of changes, signaling that a select number of dimensions are responsible for a large amount of the learning during fine-tuning. Likewise, a matrix with a large AUC learns a relatively even parameter change distribution. 

\subsection{How does few-shot learning change the parameters of the pretrained LM?}
\label{ssec:paramchange}

We first investigate how different example budgets affect parameter change. We train a few-shot \COMETTTO{} (T5) model across different example budgets ($n \in \{3, 30, 300\}$). Table~\ref{fig:attention_training_size} shows that performance increases monotonically as we grow the example budget. To explore the differences between models, we investigate the parameter changes between each fine-tuned model in Fig.~\ref{fig:attention_training_size}. We include additional heatmaps for angular change in the Appendix.





\input{figures/attention_train}

\paragraph{Findings} In Figure~\ref{fig:attention_training_size}, we observe that the $\ell_1$ distance between $\Theta^{FT}$ and $\Theta^{PT}$ increases as we increase the training budget in the decoder. 
Interestingly, we note that most parameter changes occur in the decoder of the LM (and more strongly in its later layers), rather than the encoder, implying that much of the parameter shift may be due to learning how to \textit{express} commonsense knowledge in a declarative form. 
Furthermore, when comparing the changes across parameter matrix types, the key and value matrices of the attention layers change more while the feedforward network parameters ($wi$, $wo$) remain relatively unchanged. These findings suggest the model learns how to process the structure of declarative commonsense knowledge expressions, but does not need to modify its stored representations of this knowledge learned during pretraining, hinting at a possible explanation for the success of adapters for limited-parameter finetuning of NLP models \cite{Houlsby2019ParameterEfficientTL}. 

\input{figures/sizes}

\subsection{How does model size affect knowledge model learning?}
\label{ssec:modelsize}





\input{figures/attention_model_size}

\noindent We train a few-shot \COMETTTO{} (T5) model (with $n = 30$) across different pretrained language model sizes, i.e. Small, Large, and 11B, with 60M, 770M, and 11B parameters, respectively.

\paragraph{Findings}

In Table \ref{tab:sizes}, we observe that 
the importance of model size decreases as the training budget increases, reinforcing that model scale is critically important in few-shot learning settings \cite{Brown2020LanguageMA}. 
When examining the parameter change heatmaps in Figure~\ref{fig:attention_model_size}, we note the $\ell_1$ distances between finetuned and original parameters increase as model size increases. As these distances are normalized by the number of parameters, this observation implies that when trained on the same examples, the smaller model experiences smaller magnitude gradient updates over time.  
This pattern indicates more destructive inference between parameter updates in the smaller models, an observation supported by the larger angular parameter changes in the encoders of the smaller models. 

Interestingly, several parameter matrices display high angular change but low $\ell_1$ change (e.g., in early $wo$ layers of T5-11B), suggesting that a matrix with low average absolute parameter change may still contain critical dimensions which are used to learn the task. To investigate these differences, we look at the AUC heatmaps (Fig.~\ref{fig:attention_model_size}, \emph{bottom row}), and note that as the feedforward networks get larger with increasing model size, the relative number of parameters that change to encode new information drops drastically in both the decoder and encoder. Furthermore, the lower AUCs for the encoder parameter changes indicate that the encoder layers generally experience more concentrated changes as the model size increases. These two patterns suggest that the increased capacity of larger language models allow them to encode knowledge in a less distributed manner, making their knowledge easier to access. 


\subsection{How do prompts influence knowledge model learning?}
\label{ssec:prompts}







\noindent Recent work has shown that prompts can help models elicit knowledge from pretrained language representations \cite{Feldman2019CommonsenseKM,Shin2020AutoPromptEK}. However, eliciting knowledge through zero-shot prompting has drawbacks. For example, the output is sensitive to subtle variations in the construction of the language prompt \cite{jiang20tacl}. 
Here, we explore whether prompts can accelerate few-shot learning by initializing natural language prompts for each relation (Table~\ref{table:example_templates}) training on these expressions of relations rather than initializing new relation token embeddings for each relation as in \citet{Bosselut2019COMETCT}. As a control for whether the model understands the semantics of specific relation prompts, or merely benefits from arbitrary expressions of relations using language, we also train a model where we shuffle prompts among the relations (\ie{} shuffling the \textbf{Prompt} column of Table~\ref{table:example_templates}).
%
%



\input{figures/templates}

\input{figures/attention_prompts}

\paragraph{Findings}

In Table~\ref{tab:templates}, we see that knowledge models can efficiently learn from fewer examples when relations are represented using correct natural language prompts ($\sim$10$\times$ fewer examples), an observation made by contemporaneous works on other tasks \cite{Scao2021HowMD}. 
Interestingly, however, we find that the absolute and angular distance heatmaps for prompt and embedding relation inputs are similar (Fig.~\ref{fig:attention_prompts}; $\ell_1$ heatmap in Appendix), implying that most of the parameter change during fine-tuning is not linked to the way the relation is represented. One notable difference is that models with embedding relation inputs have a significant angular change in layer 4 of the decoder $wo$ matrix. Models that use prompts see a larger angular change in layer 2 of the decoder $wo$ matrix. This discrepancy suggests that both models must learn to adapt their parameters to the new input format, but do so by adapting the same parameters at different layers. We note that there is no such concentrated angular change for the \textbf{Shuffled} setting, perhaps because the model must first \textit{unlearn} language relationships between head entities and prompts, requiring more uniform parameter updates.




%% file: figures/auc.tex
\begin{figure*}[t]
\centering
  \includegraphics[width=0.55\textwidth]{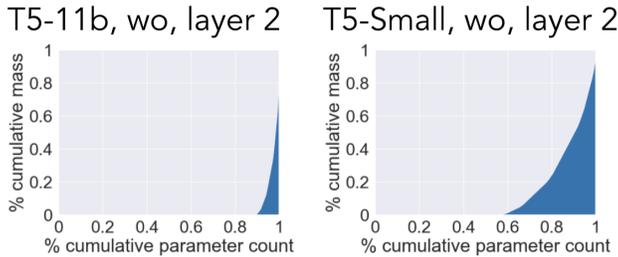}
  \caption{Example visualization of area under curve for FFN parameter matrices, $n = 30$ 
  }
  \label{fig:auc}
\end{figure*}

%% file: figures/attention_train.tex
\begin{figure*}[t]
\centering
  \includegraphics[width=\textwidth]{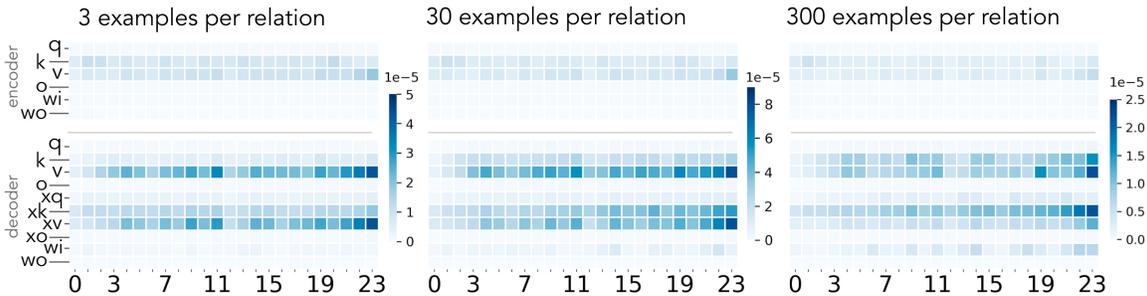}
  \caption{Normalized $\ell_1$ distance  between $\Theta^{FT}$ and $\Theta^{PT}$ for each parameter matrix at each layer of the encoder and decoder of the \COMETTTO~(T5-Large) model. We report the change for training budgets of $n=3$, $30$, and $300$. 
  }
  \label{fig:attention_training_size}
\end{figure*}

%% file: figures/sizes.tex
\begin{small}
\begin{table}[h]
\centering
\resizebox{0.7\textwidth}{!}{
    \begin{tabular}{llrrrr}
    \textbf{\# Ex} & \COMETTTO{} \textbf{Model} & \textbf{BLEU-1} & \textbf{METEOR} & \textbf{ROUGE-L} & \textbf{CIDEr} \\
    \toprule
    \multirow{3}{*}{3} & T5-Small & 13.4 $\pm$ 3.0 & 8.8 $\pm$ 3.1 & 11.5 $\pm$ 1.1 & 6.8 $\pm$ 1.0 \\
    & T5-Large & 23.5 $\pm$ 1.3 & 13.6 $\pm$ 1.1 & 19.1 $\pm$ 1.2 & 12.6 $\pm$ 2.1 \\
    & T5-11B &\textbf{24.2} $\pm$ 1.7 & \textbf{14.4} $\pm$ 1.9 & \textbf{21.3} $\pm$ 3.7 & \textbf{18.1} $\pm$ 7.9 \\
    \midrule
    \multirow{3}{*}{30} & T5-Small & 26.2 $\pm$ 2.0 & 15.5 $\pm$ 0.4 & 20.8 $\pm$ 0.6 & 11.9 $\pm$ 0.3 \\
    & T5-Large & 29.6 $\pm$ 0.6 & 16.6 $\pm$ 0.4 & 23.0 $\pm$ 0.5 & 14.3 $\pm$ 0.6 \\
    & T5-11B &\textbf{31.9} $\pm$ 0.3 & \textbf{18.7} $\pm$ 0.5 & \textbf{26.0} $\pm$ 0.6 & \textbf{19.8} $\pm$ 1.2 \\
    \midrule
    \multirow{3}{*}{300} & T5-Small & 37.0 $\pm$ 1.0 & 23.5 $\pm$ 0.4 & 35.9 $\pm$ 0.9 & 36.8 $\pm$ 2.2 \\
    & T5-Large & \textbf{39.1} $\pm$ 0.8 & 26.3 $\pm$ 0.4 & \textbf{40.4} $\pm$ 0.9 & 46.6 $\pm$ 2.0 \\
    & T5-11B &\textbf{39.1} $\pm$ 0.9 & \textbf{26.7} $\pm$ 0.9 & 40.1 $\pm$ 1.5 & \textbf{48.5} $\pm$ 2.8 \\
    \bottomrule
    
    \end{tabular}}
  \caption{Effect of model size for few-shot commmonsense knowledge modeling performance. 
  }
  \label{tab:sizes}
\end{table}
\end{small}

%% file: figures/attention_model_size.tex
\begin{figure}[t]
\centering
  \includegraphics[width=0.8\linewidth]{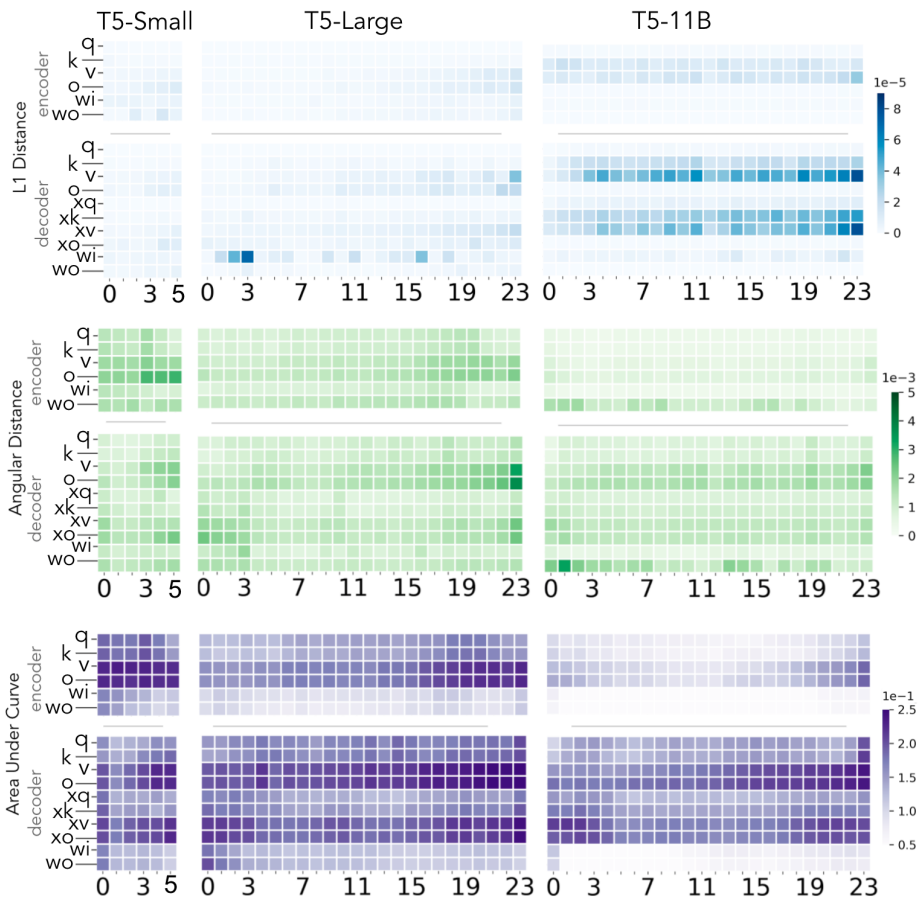}
  \caption{Parameter change measures for different knowledge model sizes. 
  }
  \label{fig:attention_model_size}
\end{figure}

%% file: figures/templates.tex

\begin{table}[t]
\centering
\resizebox{0.7\linewidth}{!}{
    \begin{tabular}{llrrrr}
    \textbf{\# Ex} & \textbf{Input} & \textbf{BLEU-1} & \textbf{METEOR} & \textbf{ROUGE-L} & \textbf{CIDEr} \\
    \toprule
    \multirow{2}{*}{3} & Prompts & \textbf{24.2} $\pm$ 1.7 & \textbf{14.4} $\pm$ 1.9 & \textbf{21.3} $\pm$ 3.7 & \textbf{18.1} $\pm$ 7.9 \\
    & Embedding & 13.9 $\pm$ 1.3 & 11.4 $\pm$ 1.2 & 13.5 $\pm$ 0.9 & 7.4 $\pm$ 0.8 \\
    \midrule
    \multirow{3}{*}{30} & Prompts & \textbf{31.9} $\pm$ 0.3 & \textbf{18.7} $\pm$ 0.5 & \textbf{26.0} $\pm$ 0.6 & \textbf{19.8} $\pm$ 1.2 \\
    & Embedding & 18.1 $\pm$ 0.8 & 13.5 $\pm$ 1.1 & 16.7 $\pm$ 1.1 & 9.7 $\pm$ 1.7 \\
    & Shuffled & 15.6 $\pm$ 0.8 & 11.3 $\pm$ 0.5 & 14.2 $\pm$ 0.5 & 8.3 $\pm$ 0.8 \\
    \bottomrule
    
    \end{tabular}}
  \caption{Prompts accelerate few-shot commonsense interface learning. 
  }
  \label{tab:templates}

\end{table}

%% file: figures/attention_prompts.tex
\begin{figure}[t]
\centering
  \includegraphics[width=0.95\linewidth]{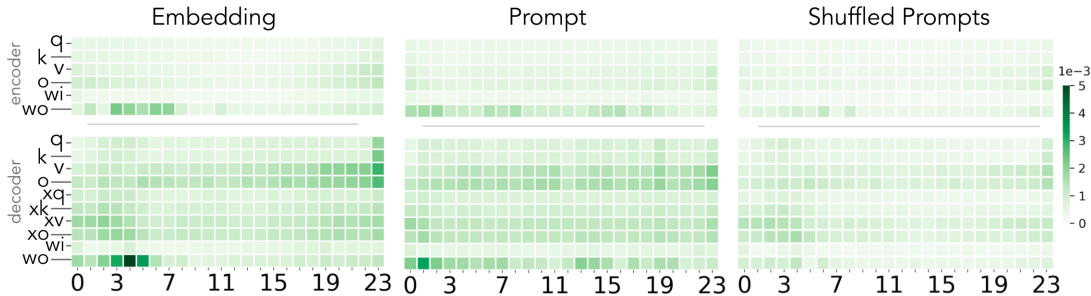}
  \caption{Average angular distance between original and finetuned parameters for prompts, shuffled prompts, and initialized relation embeddings. Each model is trained with $n = 30$. 
  }
  \label{fig:attention_prompts}
\end{figure}

%% file: sections/6-related.tex
\section{Related Work}



\paragraph{Commonsense Knowledge Models} Our work uses commonsense knowledge models, first proposed by \citet{Bosselut2019COMETCT}, to learn from commonsense knowledge graphs. \citet{Hwang2020COMETATOMIC2O} also trained commonsense models on \atomicTT, but focused on fully-supervised learning.
Other works have developed commonsense knowledge models that are grounded to visual scenes \cite{Park2020VisualCOMETRA,Da2020EditedMU}, requiring multimodal commonsense inference generation. 
Recent works extend commonsense knowledge models beyond generating single-hop inferences and generate multi-branch \cite{Bosselut2019DynamicKG} and multi-hop \cite{wang-etal-2020-connecting} inferential structures. Commonsense knowledge base completion is also a closely related task to commonsense inference generation \cite{li2016commonsense,saito2018commonsense}. Recent works on this task combine language and graph structure representations for improved generalization \cite{Malaviya2019ExploitingSA,Wang2020InductiveLO}. Our work differs from these prior studies by investigating commonsense knowledge models in few-shot settings, and analyzing the emergence of their capabilities as a function of parameter change.

\vspace{-1ex}
\paragraph{Few-shot Learning with Prompting} In recent years, the term few-shot has taken two meanings: the classical definition of training on limited examples \cite{FeiFei2006OneshotLO, NIPS2004_ef1e491a}, and a new \textit{in-context} definition where models are given examples as a prepended augmentation of their context and process these examples as input to recognize the structure of a task \cite{Brown2020LanguageMA}. Various contemporaneous works related to ours have studied the use of natural language and artificial prompts in language models \cite{Schick2020ItsNJ, Gao2020MakingPL, Tam2021ImprovingAS, Schick2021ExploitingCF} as accelerators of few-shot learning. \citet{Scao2021HowMD} further empirically study the data efficiency advantage gained from using prompting to represent a task. Our work differs from these studies in that we do not use prompts to map a classification task to masked language modeling, but instead use prompts as priming text for a generation task. Most similar to our approach is perhaps \citet{Schick2020FewShotTG}, which uses prompts for few-shot summarization, but their work does not focus on how few-shot training affects the learned representations of their model.




%% file: sections/7-conclusion.tex
\section{Conclusion}


In this work, we propose few-shot commonsense knowledge modeling, where models are finetuned on a limited number of examples to evaluate how efficiently they can adapt their pretrained representations to the task of hypothesizing commonsense knowledge tuples. Our results demonstrate that large language models require few examples to adapt their learned representations to the task and our analysis explores how their parameters change to enable this rapid emergence of commonsense representation ability.

%% file: sections/8-appendix.tex
\appendix






\section{Accuracy in zero-shot MLM setting}
\label{sec:app:t5}

\noindent While little work has explored few-shot knowledge completion, recent works have investigated performance of zero-shot knowledge graphs \cite{Petroni2019LanguageMA, Feldman2019CommonsenseKM}. Thus, we investigate the ability of T5-11B to complete commonsense knowledge in a zero-shot setting. 

\begin{table}[h]
    \centering
\begin{tabular}{lrrrr}
\textbf{Model} & \textbf{BLEU-1} & \textbf{METEOR} & \textbf{ROUGE-L} & \textbf{CIDEr} \\
\toprule
T5 - Zero-shot & 6.7 & 7.8 & 7.3 & 7.3 \\
T5 - Few-shot ($n = 3$) & 31.9 & 18.7 & 26.0 & 19.8 \\
T5 - Fully-supervised & 48.2 & 34.1 & 50.0 & 66.4 \\
\bottomrule
\end{tabular}
\caption{Zero-shot performance of T5-11B}
\label{tab:t5-zero}
\end{table}

\noindent We use prompts to leverage the masking objective of the language model pretraining. Since the mask only predicts several tokens at a time, for relations with longer length tail entities, we allow the model to predict up to 7 mask tokens in succession, or until the model predicts an empty string for the mask. We suggest that this is still only a workaround, and masked models are poor predictors of longer length tail entities, as indicated by our results above.

\vspace{1ex}
\section{Additional Results on the Effect of Relation Input Format}

In Table~\ref{tab:templates_all}, we provide further experimental results on the effect of relation input format on few-shot performance. Namely, we extend the results from Table~\ref{tab:templates} with the case $n=300$. These results confirm that knowledge models can efficiently learn from fewer examples when relations are represented using natural language prompts. We also show the results from using paraphrases of the main prompts for training (original and paraphrased prompts can be found in Tables~\ref{table:example_templates_large} and \ref{table:example_templates_paraphrase}, respectively). We find that the paraphrased prompts do cause a slight drop in performance, but that this drop is generally close to the margin of error, indicating that while prompt formulation is an important consideration \cite{jiang20tacl}, fine-tuning on the prompts does make the model less sensitive to prompt variations.

\begin{table}[h]
\centering
\resizebox{0.7\linewidth}{!}{
    \begin{tabular}{llrrrr}
    \textbf{\# Ex} & \textbf{Input} & \textbf{BLEU-1} & \textbf{METEOR} & \textbf{ROUGE-L} & \textbf{CIDEr} \\
    \toprule
    \multirow{2}{*}{3} & Prompts & \textbf{24.2} $\pm$ 1.7 & \textbf{14.4} $\pm$ 1.9 & \textbf{21.3} $\pm$ 3.7 & \textbf{18.1} $\pm$ 7.9 \\
    & Embedding & 13.9 $\pm$ 1.3 & 11.4 $\pm$ 1.2 & 13.5 $\pm$ 0.9 & 7.4 $\pm$ 0.8 \\
    \midrule
    \multirow{4}{*}{30} & Prompts & \textbf{31.9} $\pm$ 0.3 & \textbf{18.7} $\pm$ 0.5 & \textbf{26.0} $\pm$ 0.6 & \textbf{19.8} $\pm$ 1.2 \\
    & Embedding & 18.1 $\pm$ 0.8 & 13.5 $\pm$ 1.1 & 16.7 $\pm$ 1.1 & 9.7 $\pm$ 1.7 \\
    & Shuffled Prompts & 15.6 $\pm$ 0.8 & 11.3 $\pm$ 0.5 & 14.2 $\pm$ 0.5 & 8.3 $\pm$ 0.8 \\
     & Paraphrased & 30.5 $\pm$ 1.3 & 18.3 $\pm$ 0.5 & 24.5 $\pm$ 0.8 & 18.4 $\pm$ 1.4 \\
    \midrule
    \multirow{2}{*}{300} & Prompts & \textbf{39.1} $\pm$ 0.9 & \textbf{26.7} $\pm$ 0.9 & \textbf{40.1} $\pm$ 1.5 & \textbf{48.5} $\pm$ 2.8 \\
    & Embedding & 25.2 $\pm$ 0.8 & 18.5 $\pm$ 0.9 & 26.4 $\pm$ 1.9 & 26.8 $\pm$ 4.3 \\
    \bottomrule
    
    \end{tabular}}
  \caption{Effect of relation input format on few-shot performance. Prompts accelerate few-shot commonsense interface learning. We show mean performance over 5 random splits of training examples, and standard deviation ($\pm$) between splits.}
  \label{tab:templates_all}

\end{table}

    


\begin{table}[]
\centering
\begin{tabular}{ll}
\textbf{Relation} & \textbf{Template}                          \\ \toprule
ObjectUse         & \{\} is used for                           \\
AtLocation        & You are likely to find \{\} in             \\
MadeUpOf          & \{\} is made up of                         \\
HasProperty       & \{\} is                                    \\
CapableOf         & \{\} can                                   \\
Desires           & \{\} wants                                 \\
NotDesires        & \{\} does not want                         \\
isAfter           & Something that happens after \{\} is       \\
HasSubEvent       & Something you might do while \{\} is       \\
isBefore          & Something that happens before \{\} is      \\
HinderedBy        & \{\} is hindered by                        \\
Causes            & Sometimes \{\} causes                      \\
xReason           & \{\}. The reason for PersonX doing this is \\
isFilledBy        & \{\} can be filled by                      \\
xNeed             & But before \{\}, PersonX needed            \\
xAttr             & \{\} is seen as                            \\
xEffect           & As a result of \{\}, PersonX will          \\
xReact            & As a result of \{\}, PersonX feels         \\
xWant             & After \{\}, PersonX would want             \\
xIntent           & Because of \{\}, PersonX wanted            \\
oEffect           & as a result of \{\}, others will           \\
oReact            & as a result of \{\}, others would feel     \\
oWant             & as a result of \{\}, others would want    \\
\bottomrule
\end{tabular}
\caption{Prompts used for relations in ATOMIC2020.}
\label{table:example_templates_large}
\end{table}

\begin{table}[]
\centering
\begin{tabular}{l}
Something you might do while \{\} is \_\_\_\\
Something you might do while design software is	determine deliverables \\
Something you might do while scuba dive is take off scuba gear \\
Something you might do while play ball is put on mitt \\
\end{tabular}
\caption{Example of augmentation experiments, following \cite{Brown2020LanguageMA}. \{\} indicates the location of the head, and the language model is asked to complete the tail by finishing the sentence.}
\label{table:example_templates_gpt3}
\end{table}

\begin{table}
\centering
\begin{tabular}{ll}
\textbf{Relation}    & \textbf{Template}                             \\
\toprule
ObjectUse   & a \{\} can be used for               \\
AtLocation  & You could find \{\} in the location  \\
MadeUpOf    & \{\} is made up of                   \\
HasProperty & \{\} will have                       \\
CapableOf   & \{\} is capable of                   \\
Desires     & a \{\} desires                       \\
NotDesires  & a \{\} does not desire               \\
isAfter     & Before \{\},                         \\
HasSubEvent & You might do \{\} while doing        \\
isBefore    & After \{\},                          \\
HinderedBy  & \{\}. This is hindered by            \\
Causes      & Sometimes \{\} causes                \\
xReason     & \{\}. PersonX did this because       \\
isFilledBy  & \{\} is filled                       \\
xNeed       & Before \{\}, PersonX needs to        \\
xAttr       & \{\}. An attribute of PersonX is     \\
xEffect     & The effect of \{\} PersonX will be   \\
xReact      & As a result of \{\}. PersonX will be \\
xWant       & After \{\}, PersonX will want to     \\
xIntent     & For \{\}, PersonX did this to        \\
oEffect     & An effect of \{\} on others will be  \\
oReact      & As a result of \{\}, other feel      \\
oWant       & After \{\}, others will want to     \\
\bottomrule
\end{tabular}
\caption{Paraphrased version of the prompts used (for paraphrased experiment in the Appendix.)}
\label{table:example_templates_paraphrase}
\end{table}

\vspace{2ex}
\newpage
\section{Additional Experiments on Parameter Change Measures}

\begin{figure}[h]
\centering
  \includegraphics[width=\linewidth]{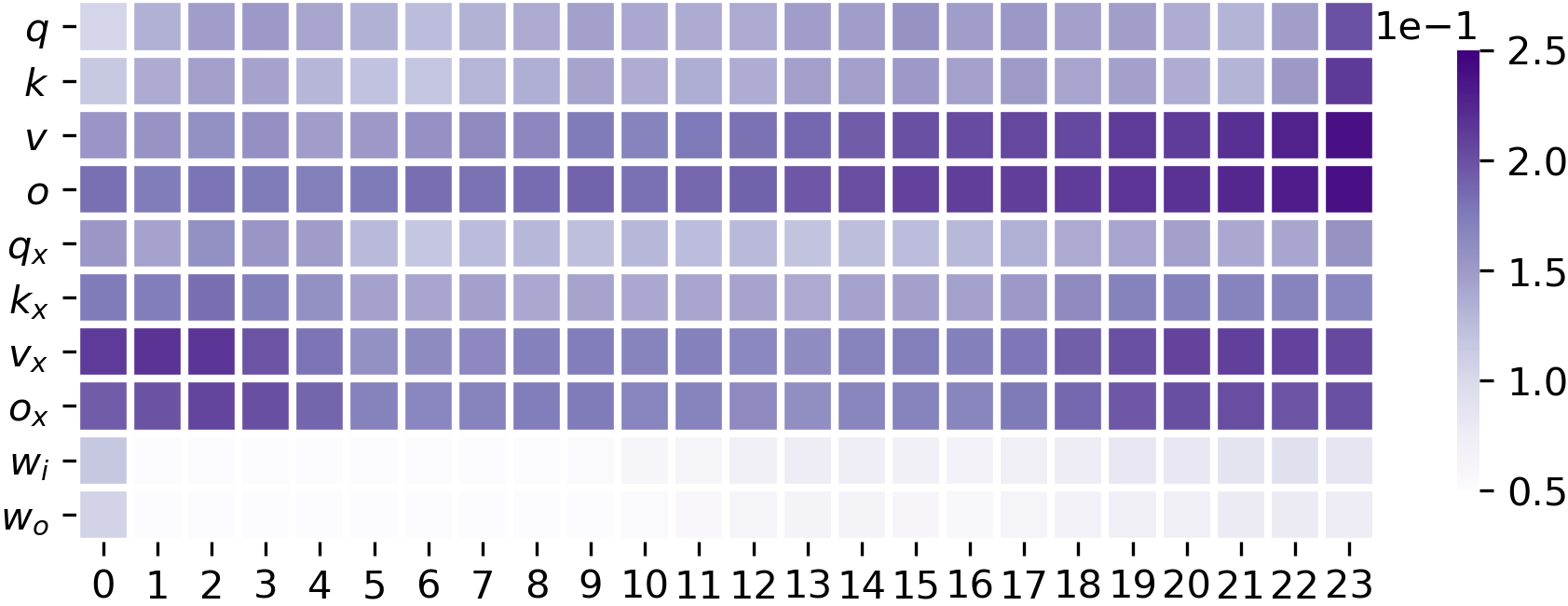}
  \caption{Area Under Curve, Decoder, T5-11B, $n = 30$}
  \label{fig:auc_1}
\end{figure}

\begin{figure}[t]
\centering
  \includegraphics[width=\linewidth]{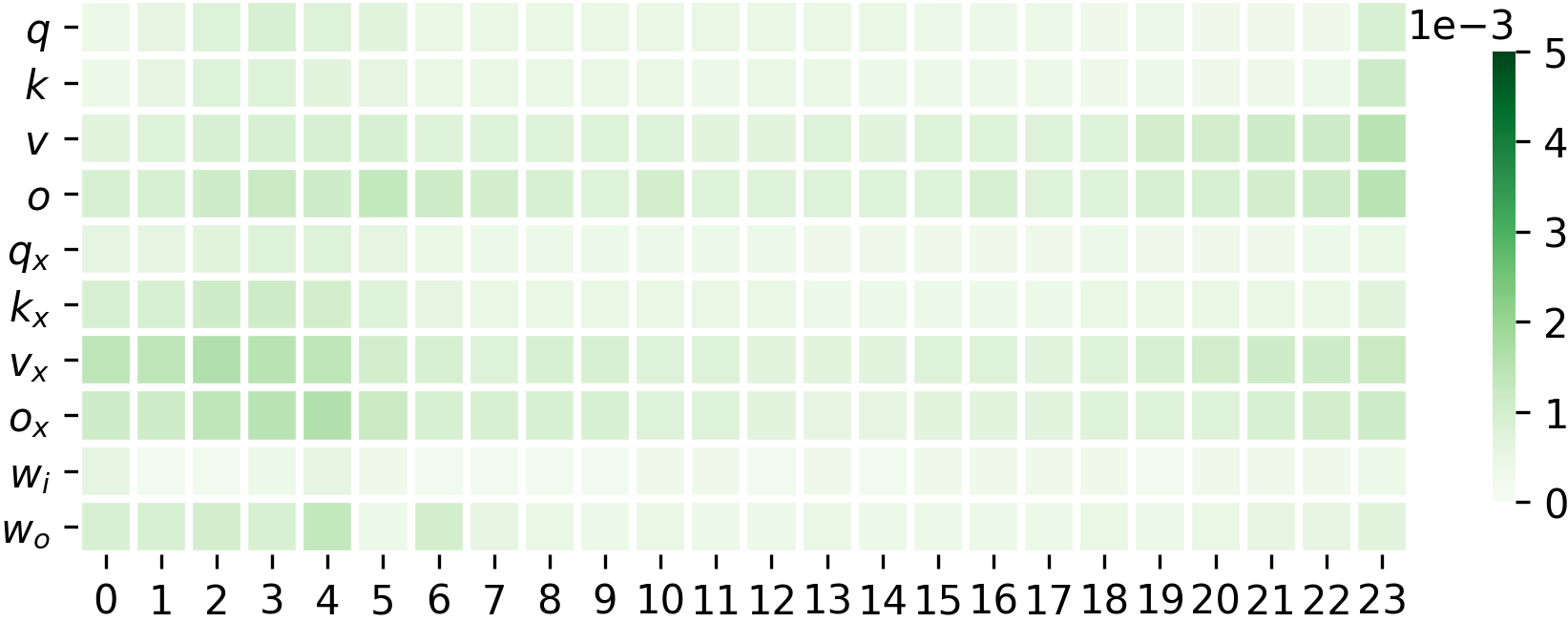}
  \caption{Angular change, Decoder, T5-11B, Shuffled Prompts, $n = 30$}
  \label{fig:ang_3}
\end{figure}

In addition, we provide extensive results of the $\ell_1$ and angular distances, as well as the distributional parameter change metric (AUC), for both the encoder and decoder of various model sizes (Small, Large, and 11B) and different example budget ($n \in \{3, 30, 300\}$) (Figures~\ref{fig:auc_1}-\ref{fig:ang_3}). When computing AUC diagrams, we round each weight change to the nearest $10^{-5}$.

\begin{figure}[h]
\centering
  \includegraphics[width=\linewidth]{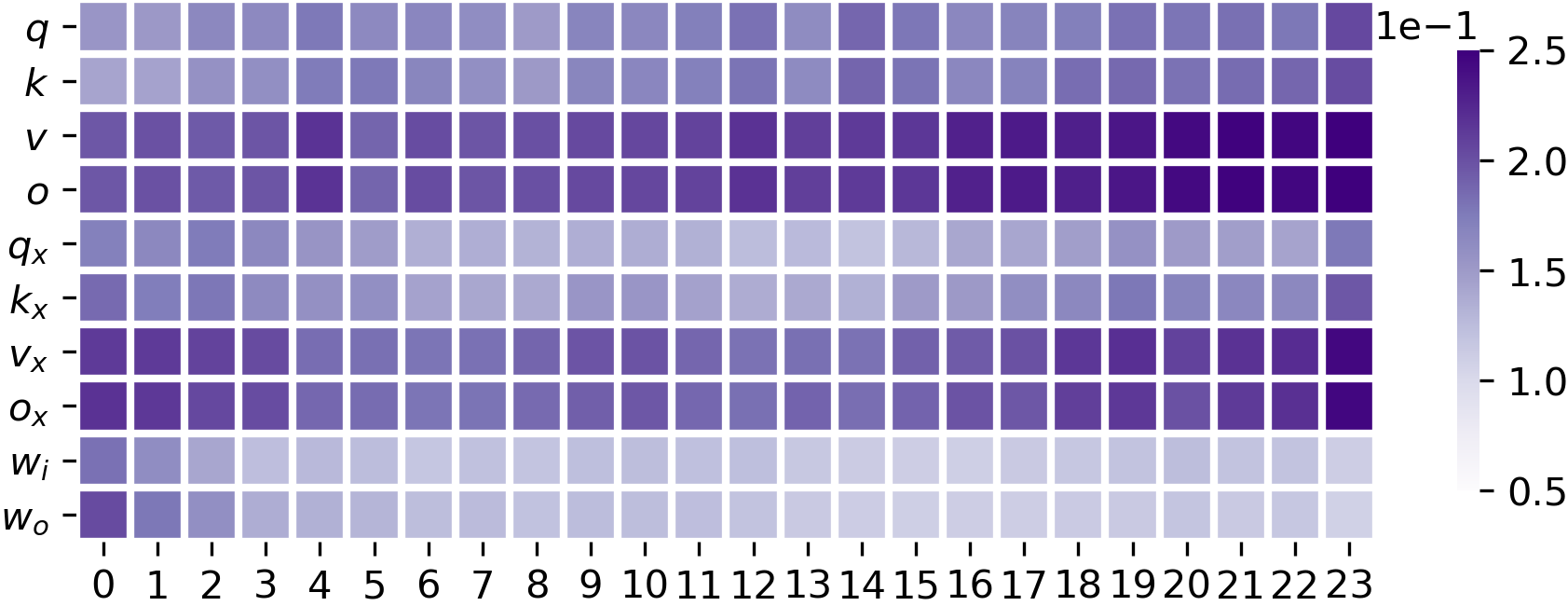}
  \caption{Area Under Curve, Decoder, T5-Large, $n = 30$}
\end{figure}

\begin{figure}[t]
\centering
  \includegraphics[width=\linewidth]{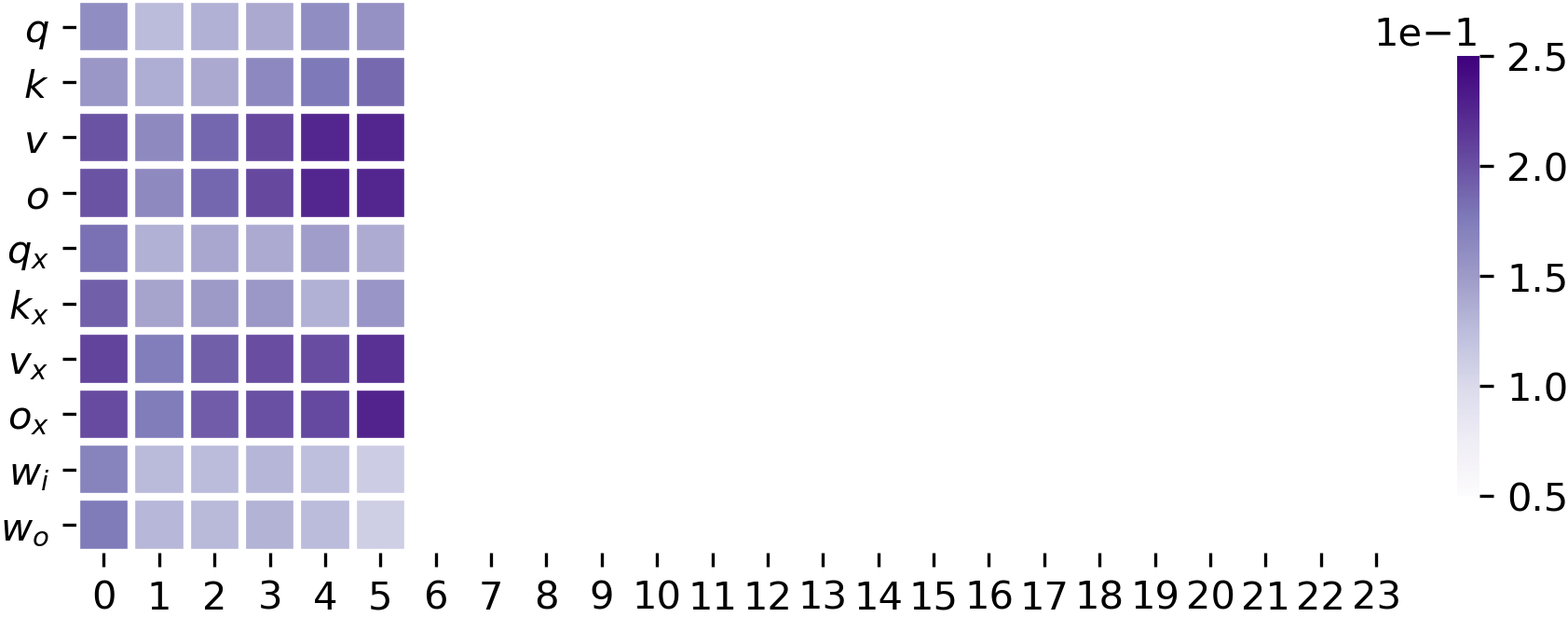}
  \caption{Area Under Curve, Decoder, T5-Small, $n = 30$}
\end{figure}

\begin{figure}[t]
\centering
  \includegraphics[width=\linewidth]{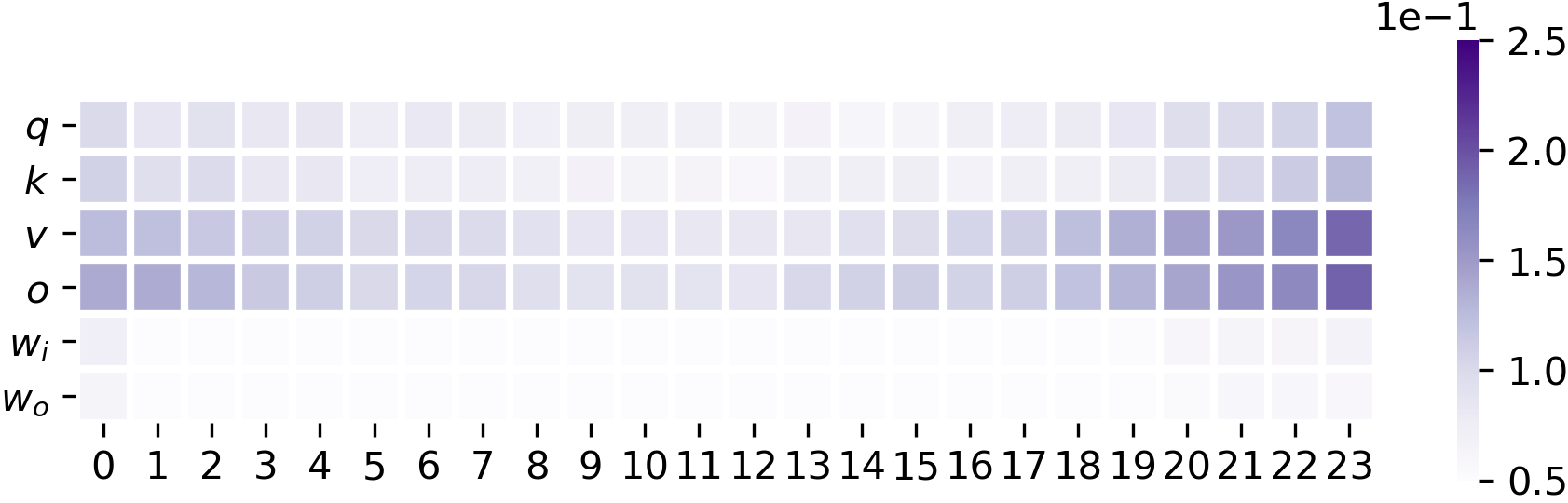}
  \caption{Area Under Curve, Encoder, T5-11B, $n = 30$}
\end{figure}

\begin{figure}[t]
\centering
  \includegraphics[width=\linewidth]{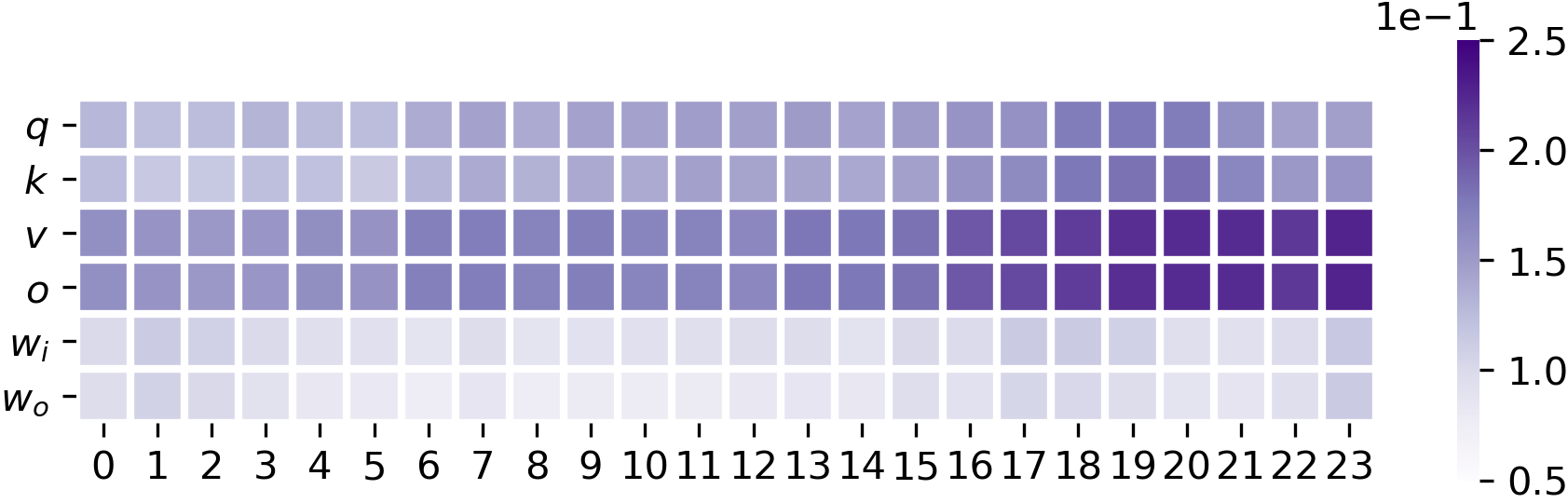}
  \caption{Area Under Curve, Encoder, T5-Large, $n = 30$}
\end{figure}

\begin{figure}[t]
\centering
  \includegraphics[width=\linewidth]{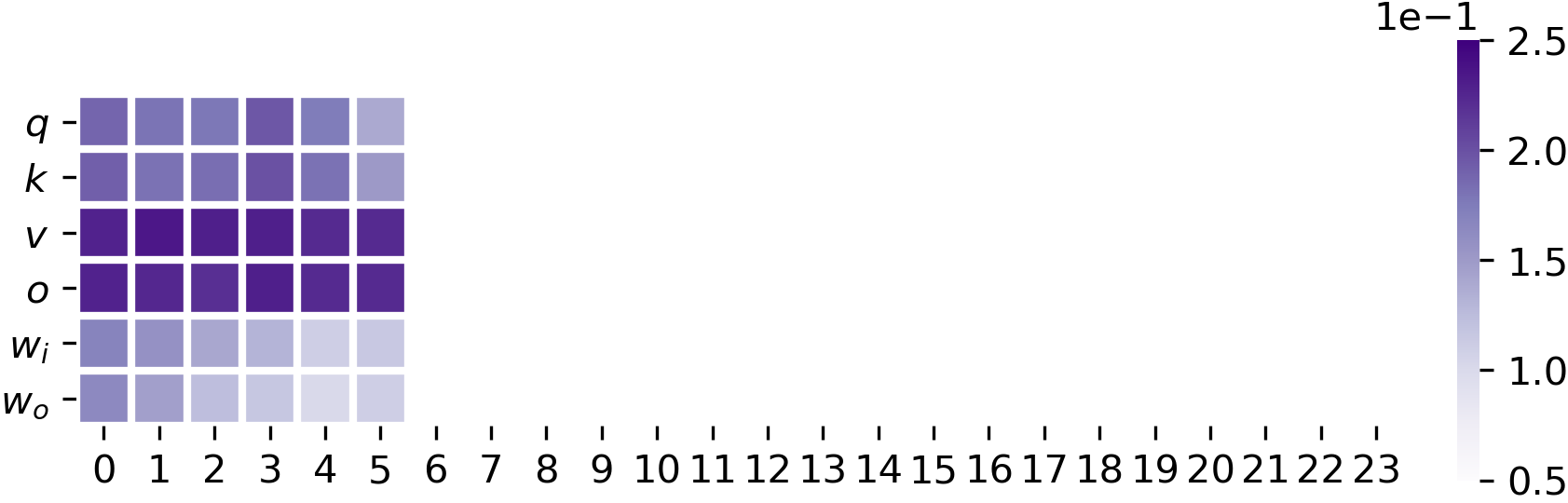}
  \caption{Area Under Curve, Encoder, T5-Small, $n = 30$}
\end{figure}

\begin{figure}[t]
\centering
  \includegraphics[width=\linewidth]{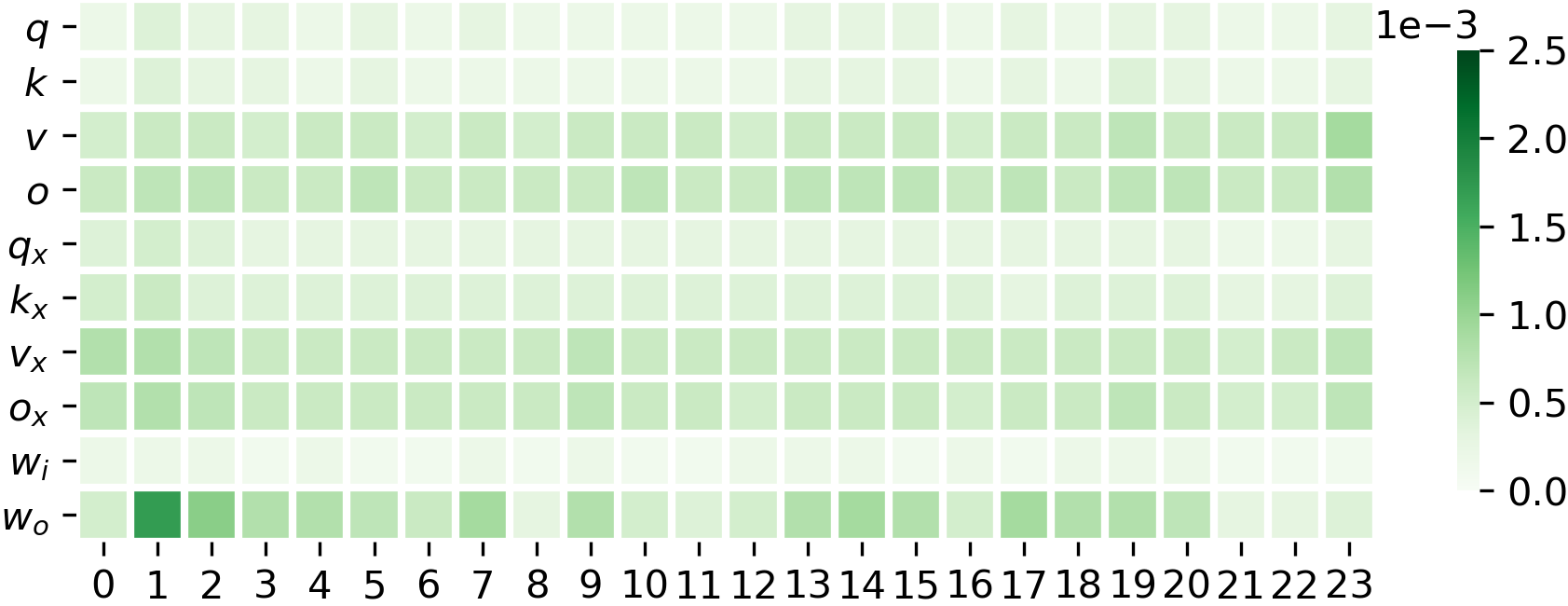}
  \caption{Angular change, Decoder, T5-11B, $n = 3$}
\end{figure}

\begin{figure}[t]
\centering
  \includegraphics[width=\linewidth]{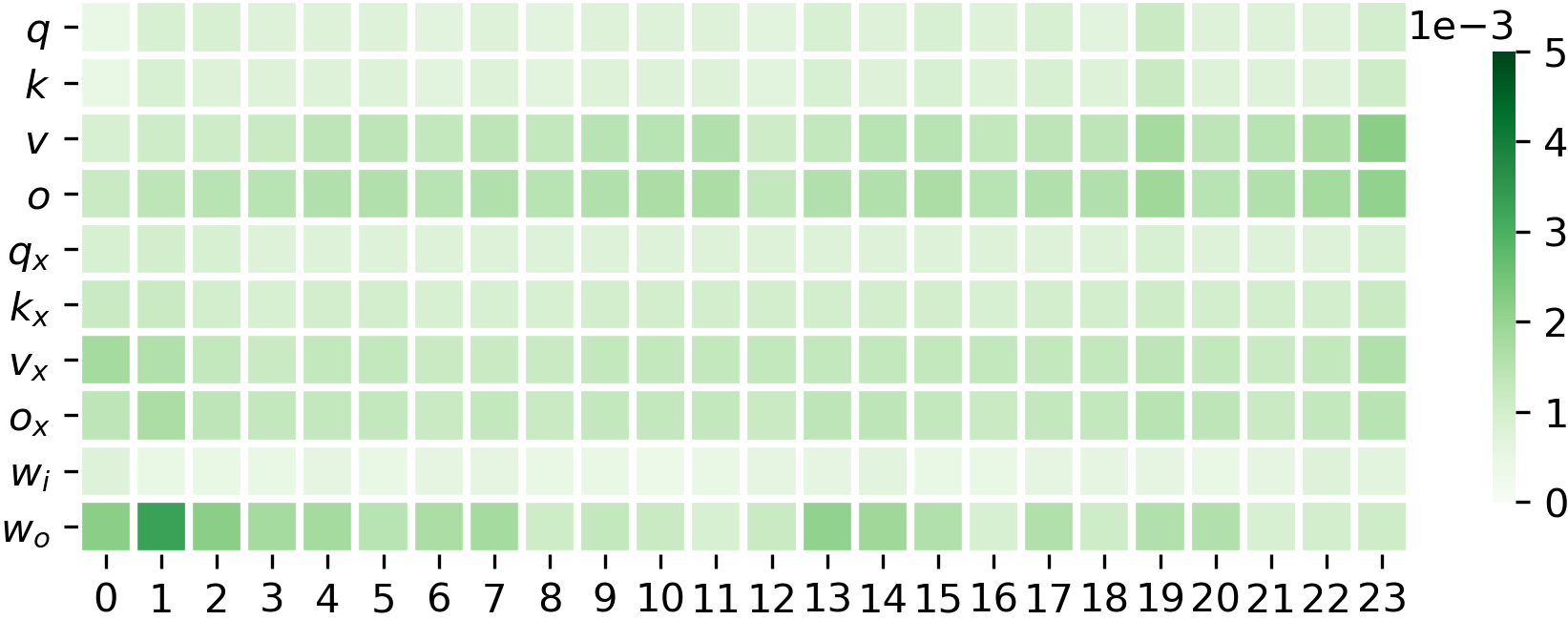}
  \caption{Angular change, Decoder, T5-11B, $n = 30$}
\end{figure}

\begin{figure}[t]
\centering
  \includegraphics[width=\linewidth]{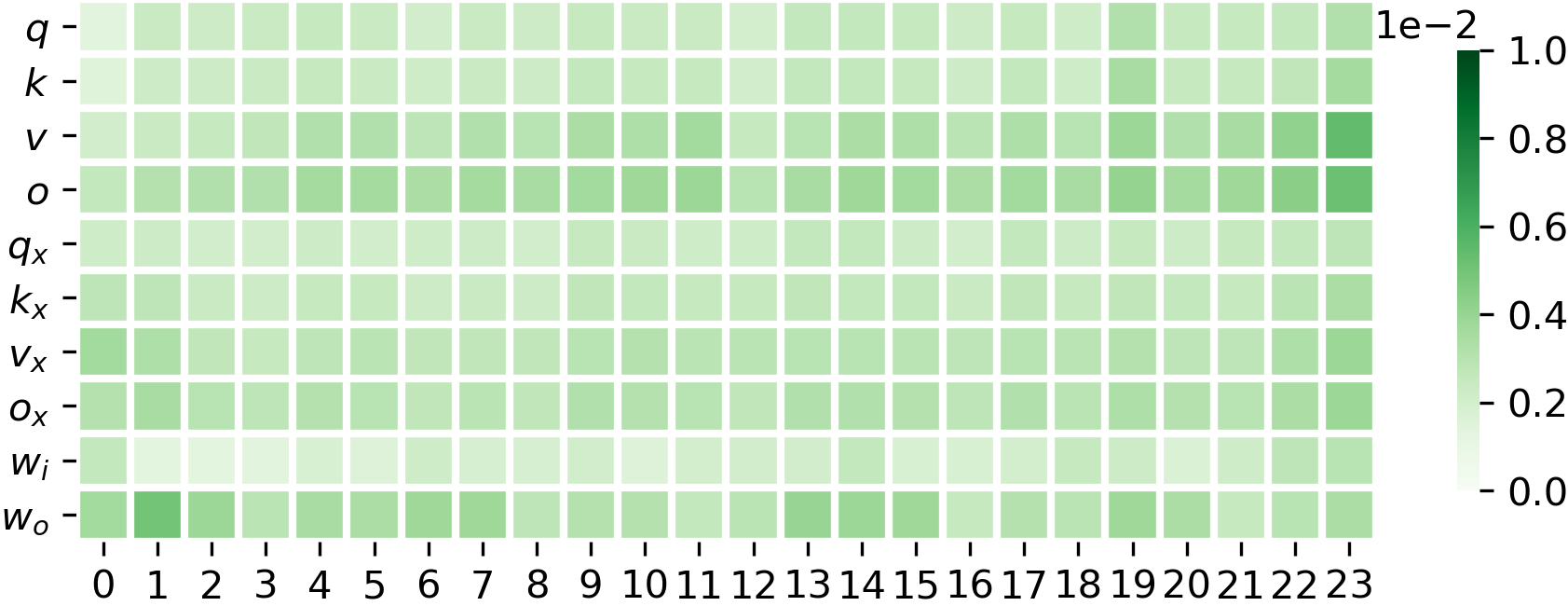}
  \caption{Angular change, Decoder, T5-11B, $n = 300$}
\end{figure}

\begin{figure}[t]
\centering
  \includegraphics[width=\linewidth]{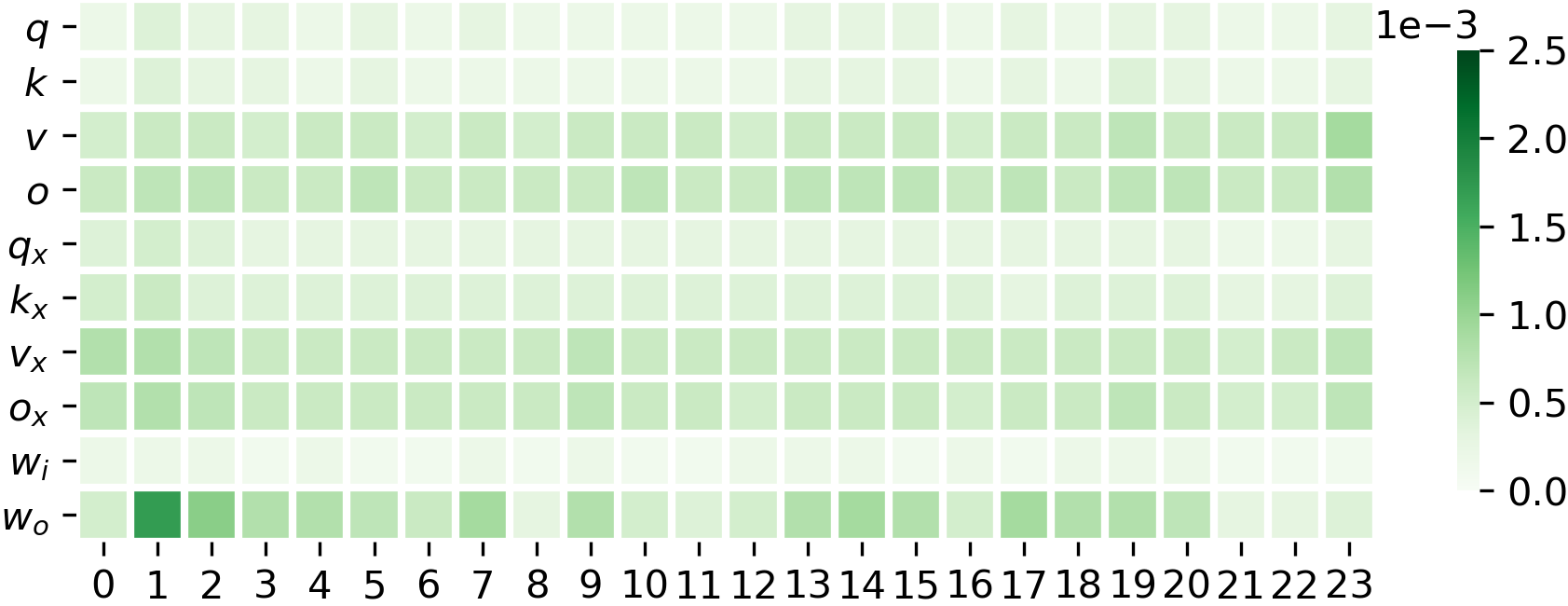}
  \caption{Angular change, Encoder, T5-11B, $n = 3$}
\end{figure}

\begin{figure}[t]
\centering
  \includegraphics[width=\linewidth]{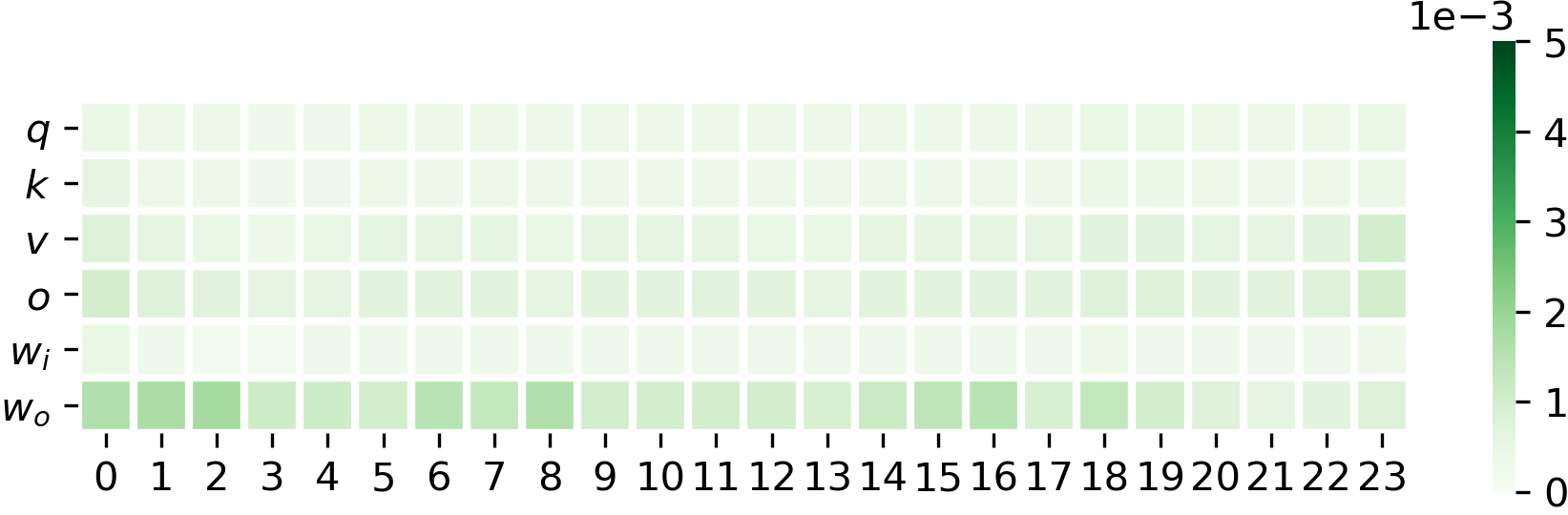}
  \caption{Angular change, Encoder, T5-11B, $n = 30$}
\end{figure}

\begin{figure}[t]
\centering
  \includegraphics[width=\linewidth]{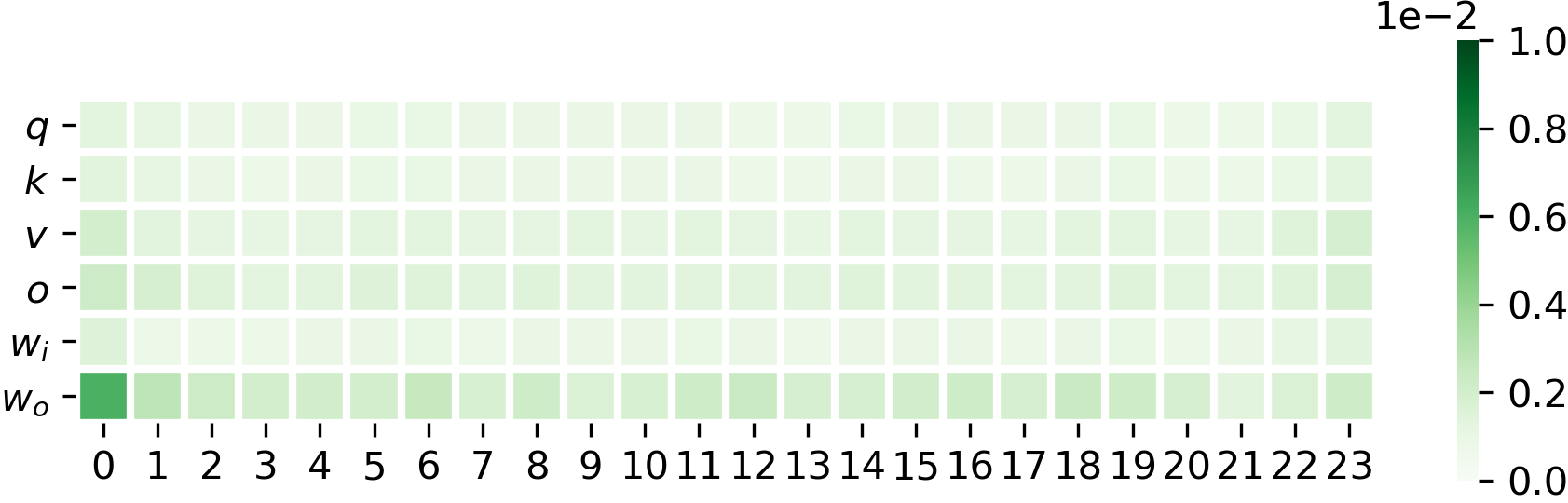}
  \caption{Angular change, Encoder, T5-11B, $n = 300$}
\end{figure}

\begin{figure}[t]
\centering
  \includegraphics[width=\linewidth]{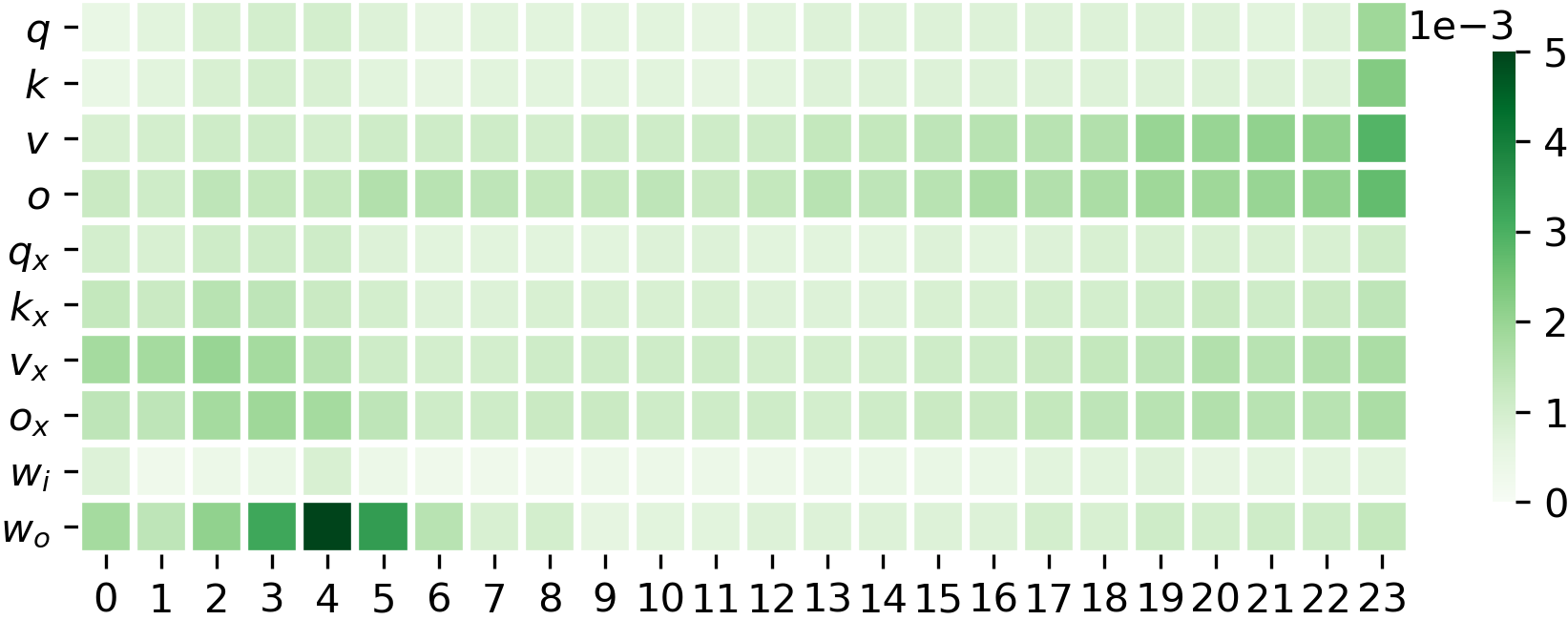}
  \caption{Angular change, Decoder, T5-11B (relation embedding), $n = 30$}
\end{figure}

\begin{figure}[t]
\centering
  \includegraphics[width=\linewidth]{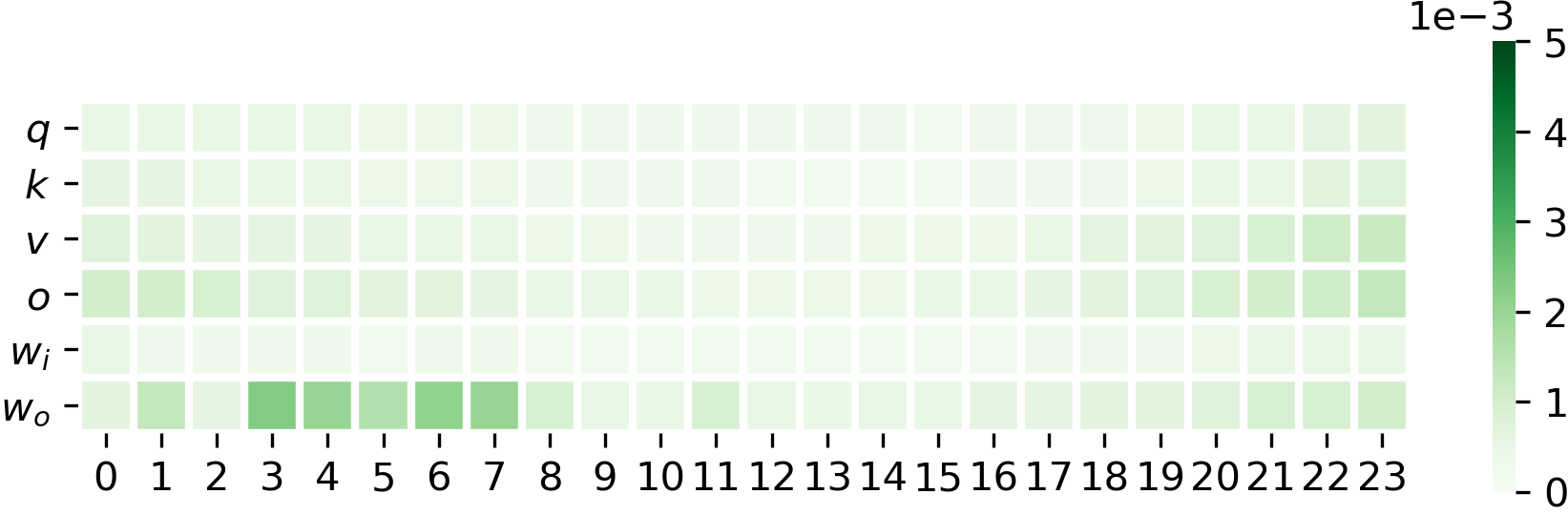}
  \caption{Angular change, Encoder, T5-11B (relation embedding), $n = 30$}
\end{figure}

\begin{figure}[t]
\centering
  \includegraphics[width=\linewidth]{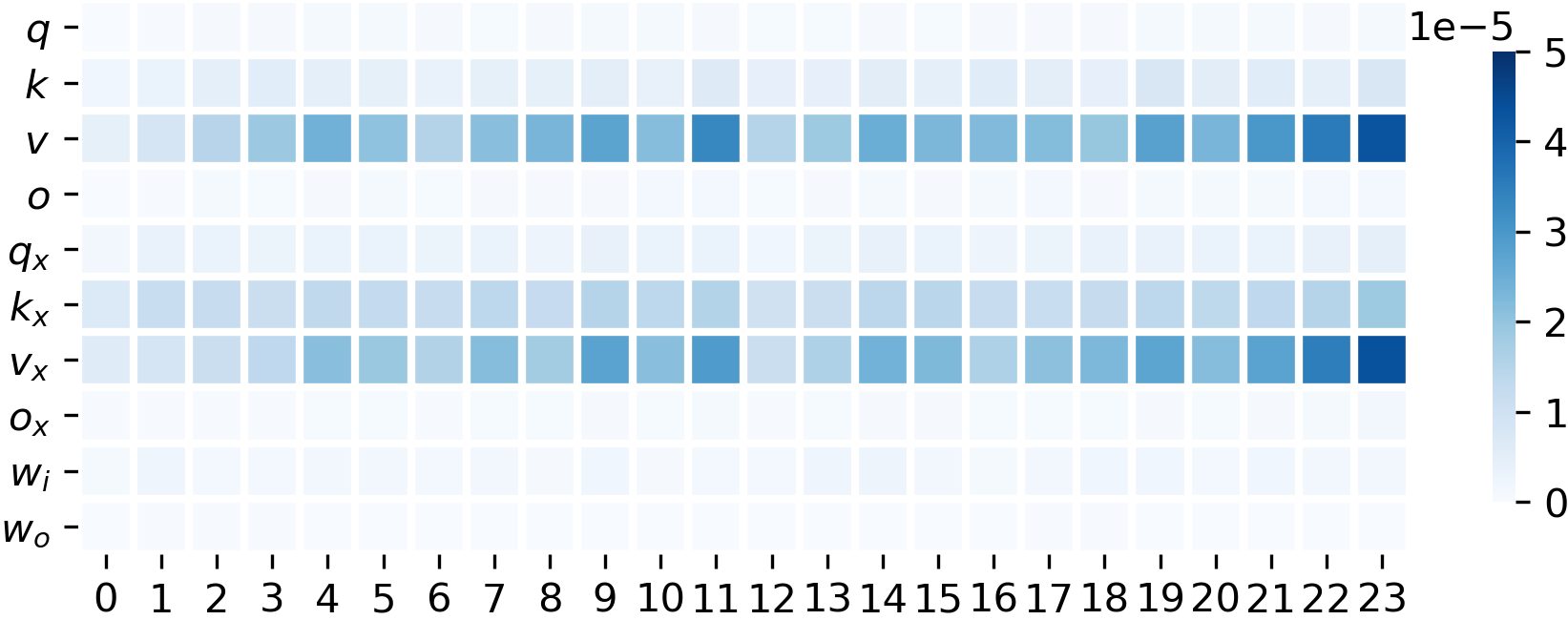}
  \caption{L1 change, Decoder, T5-11B, $n = 3$}
\end{figure}

\begin{figure}[t]
\centering
  \includegraphics[width=\linewidth]{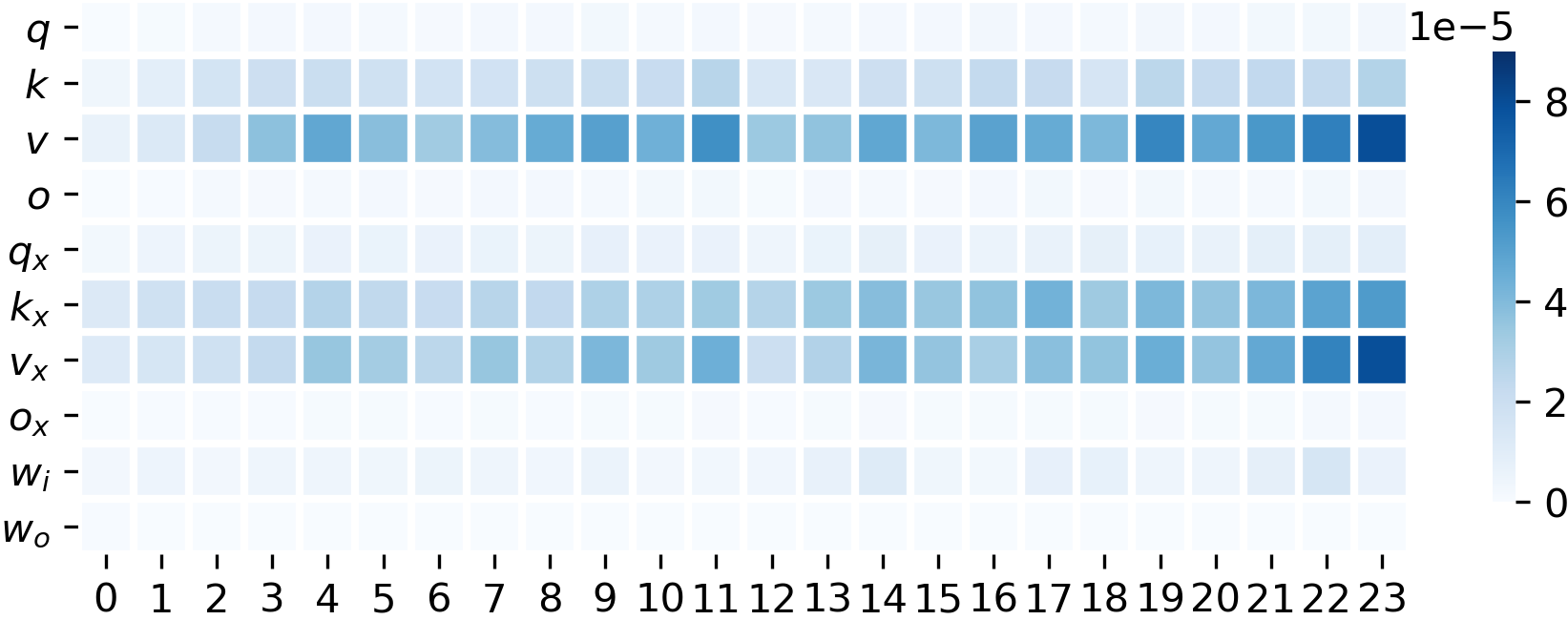}
  \caption{L1 change, Decoder, T5-11B, $n = 30$}
\end{figure}

\begin{figure}[t]
\centering
  \includegraphics[width=\linewidth]{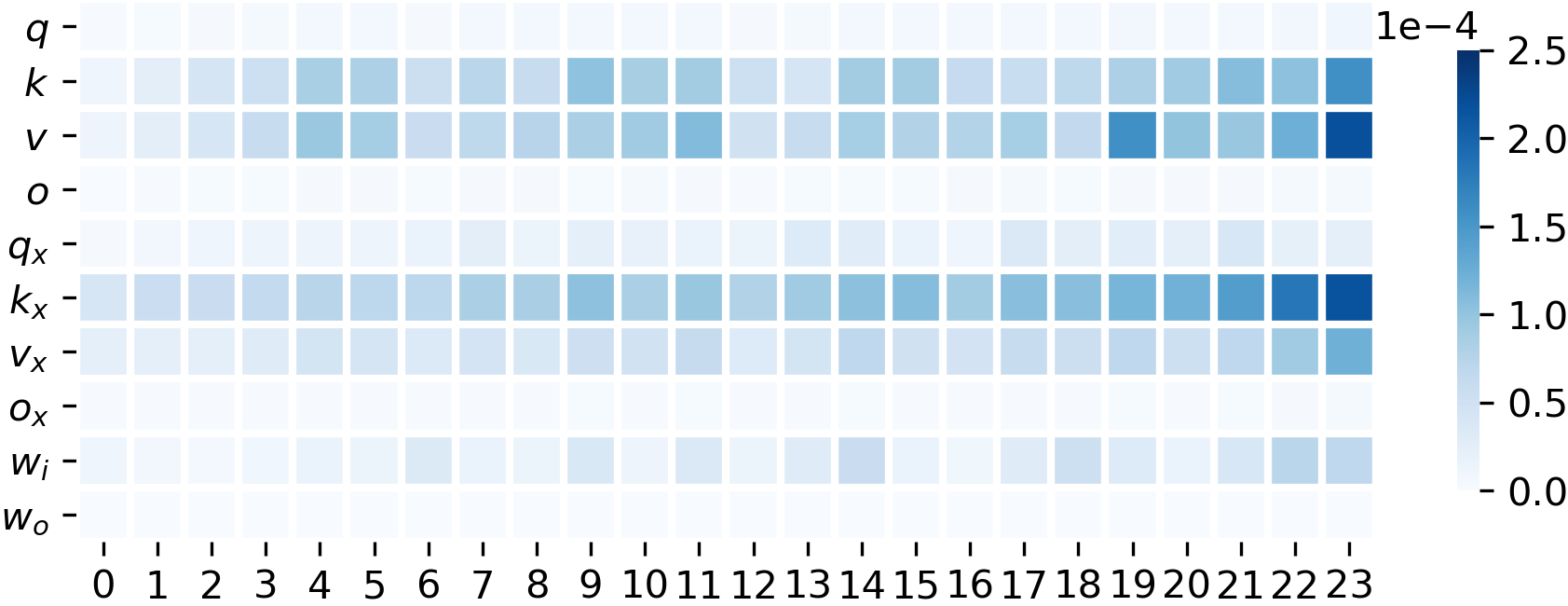}
  \caption{L1 change, Decoder, T5-11B, $n = 300$}
\end{figure}

\begin{figure}[t]
\centering
  \includegraphics[width=\linewidth]{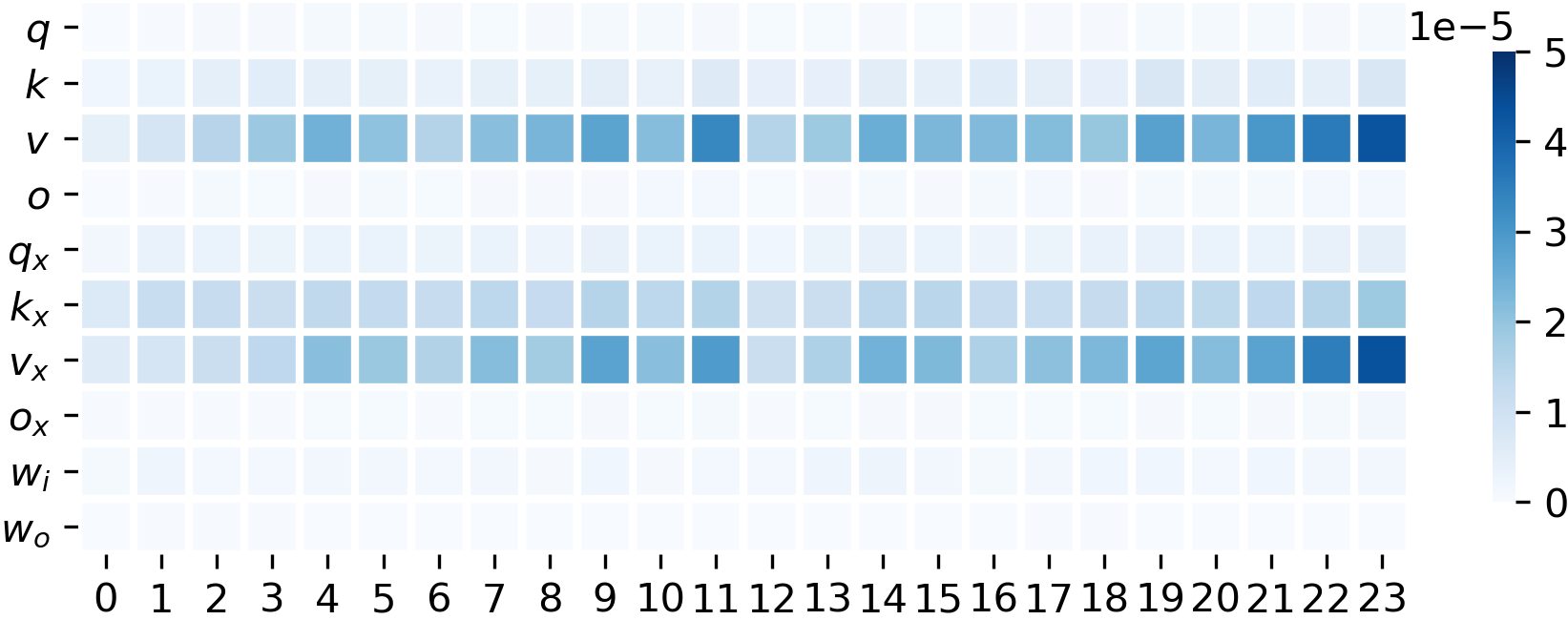}
  \caption{L1 change, Encoder, T5-11B, $n = 3$}
\end{figure}

\begin{figure}[t]
\centering
  \includegraphics[width=\linewidth]{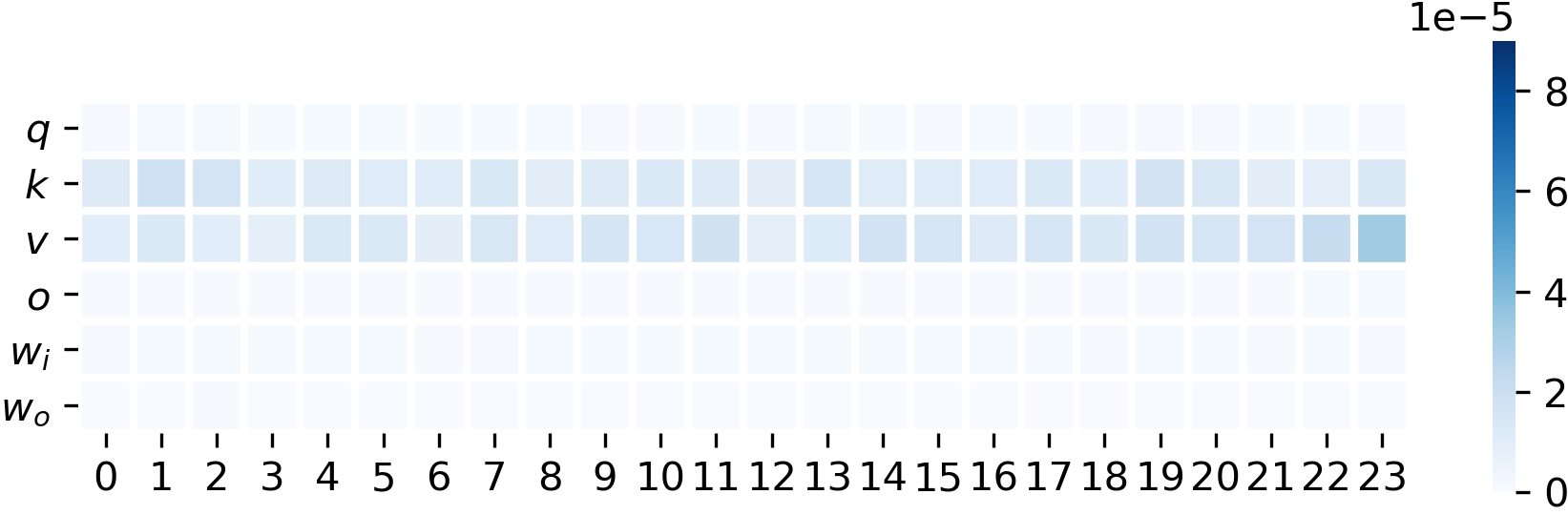}
  \caption{L1 change, Encoder, T5-11B, $n = 30$}
\end{figure}

\begin{figure}[t]
\centering
  \includegraphics[width=\linewidth]{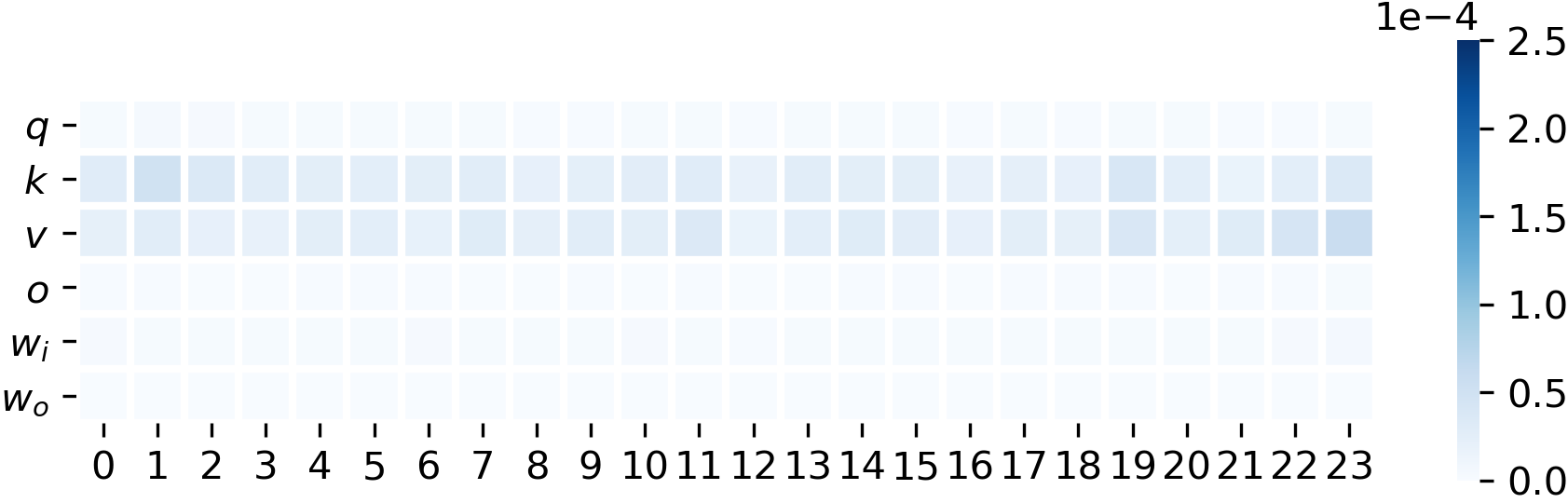}
  \caption{L1 change, Encoder, T5-11B, $n = 300$}
\end{figure}

\begin{figure}[t]
\centering
  \includegraphics[width=\linewidth]{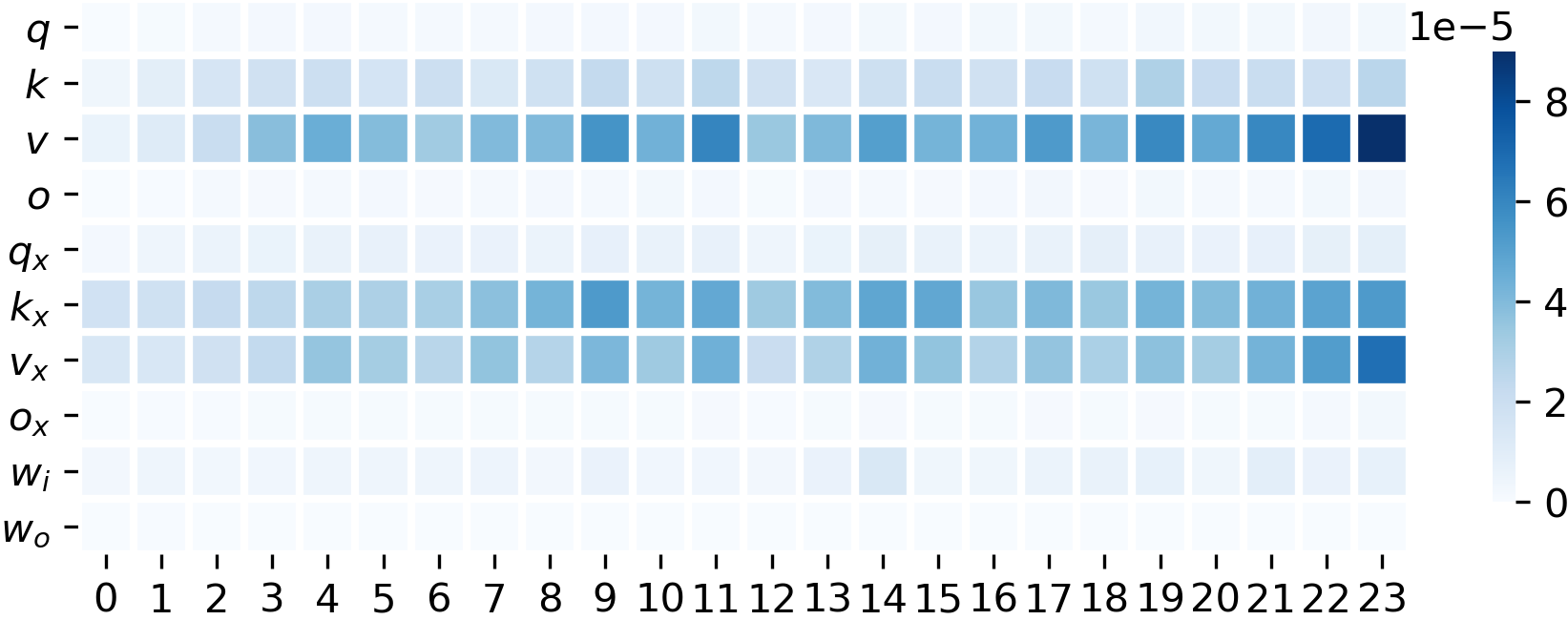}
  \caption{L1 change, Decoder, T5-11B (relation embedding), $n = 30$}
\end{figure}

\begin{figure}[t]
\centering
  \includegraphics[width=\linewidth]{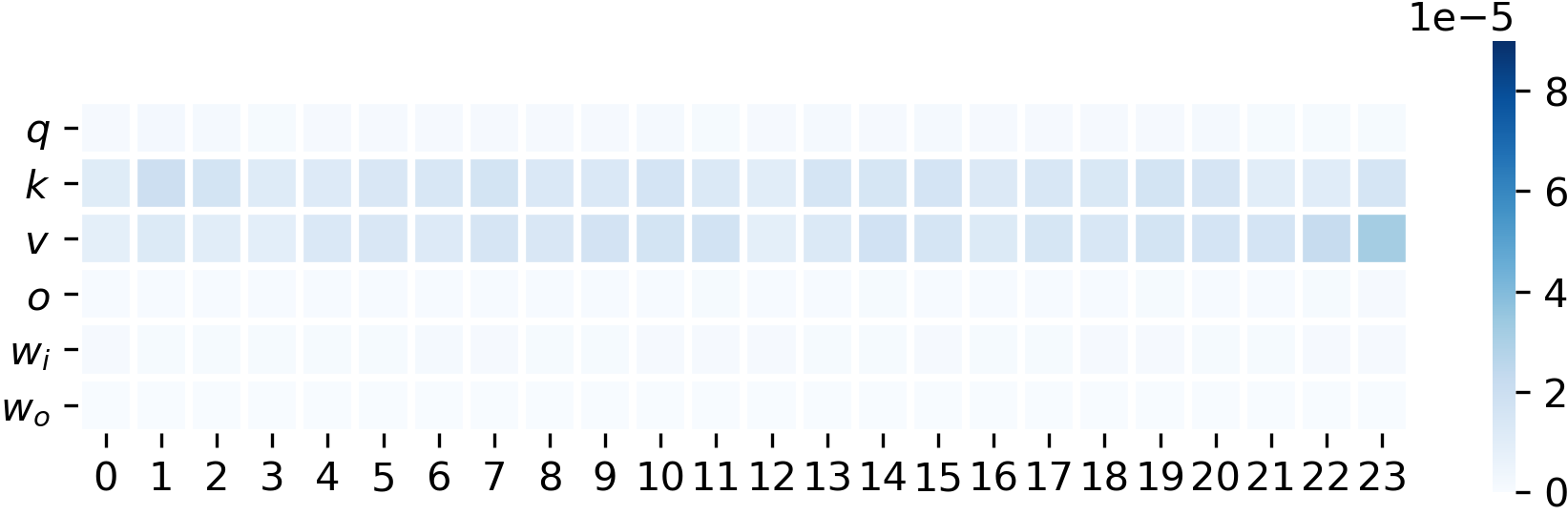}
  \caption{L1 change, Encoder, T5-11B (relation embedding), $n = 30$}
\end{figure}

\begin{figure}[t]
\centering
  \includegraphics[width=\linewidth]{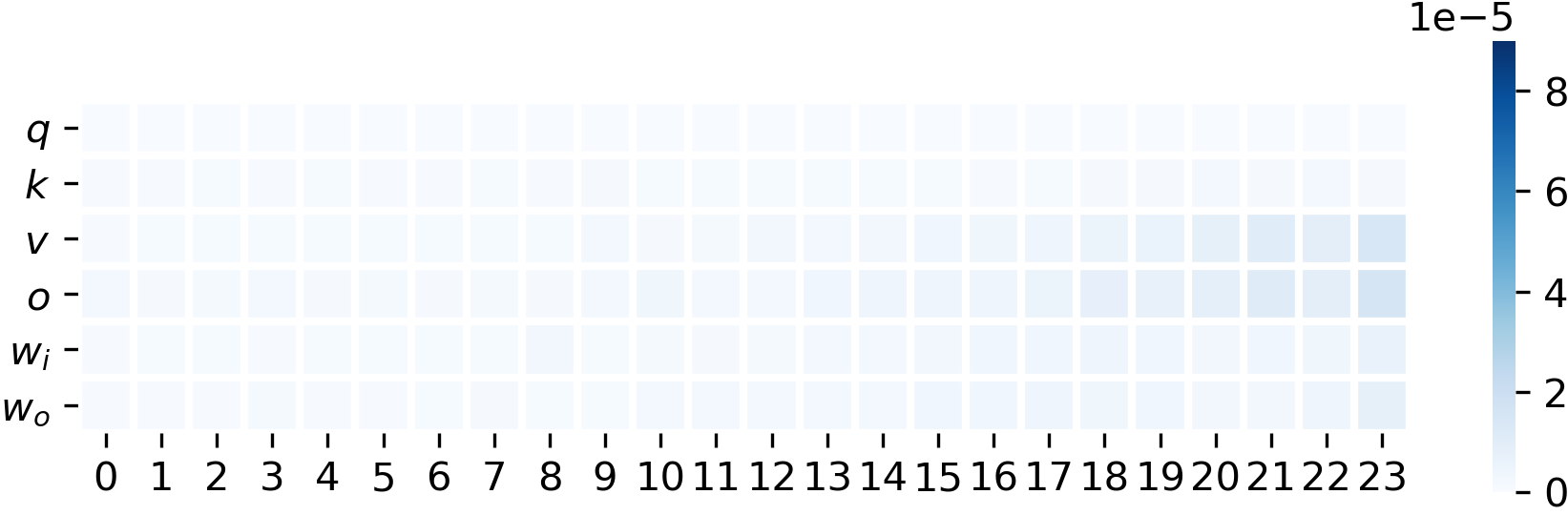}
  \caption{L1 change, Encoder, T5-Large, $n = 30$}
\end{figure}

\begin{figure}[t]
\centering
  \includegraphics[width=\linewidth]{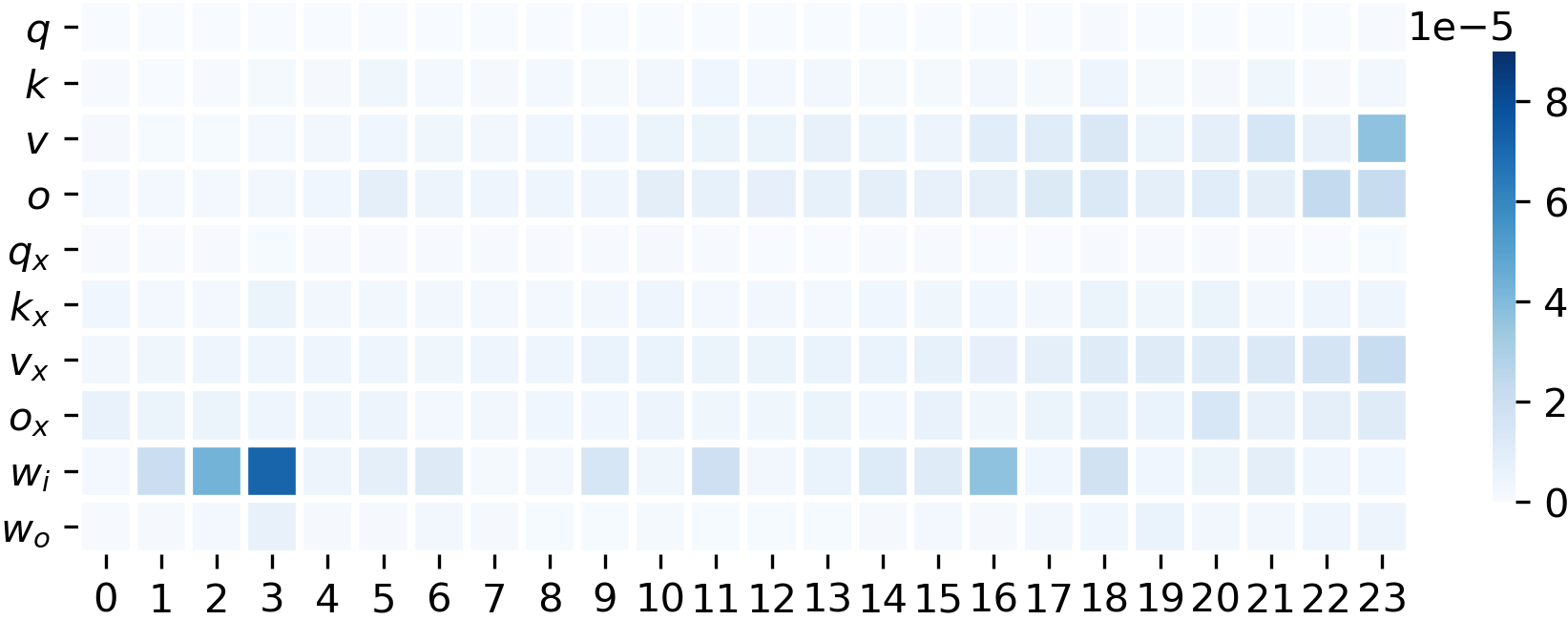}
  \caption{L1 change, Decoder, T5-Large, $n = 30$}
\end{figure}

\begin{figure}[t]
\centering
  \includegraphics[width=\linewidth]{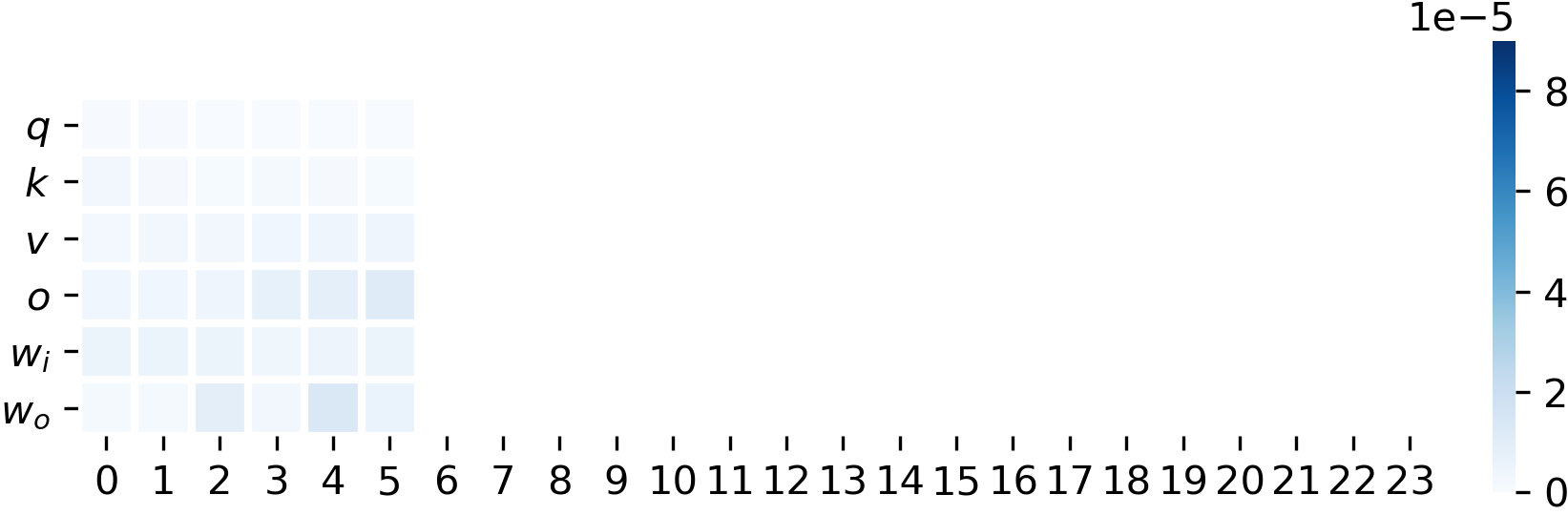}
  \caption{L1 change, Encoder, T5-Small, $n = 30$}
\end{figure}

\begin{figure}[t]
\centering
  \includegraphics[width=\linewidth]{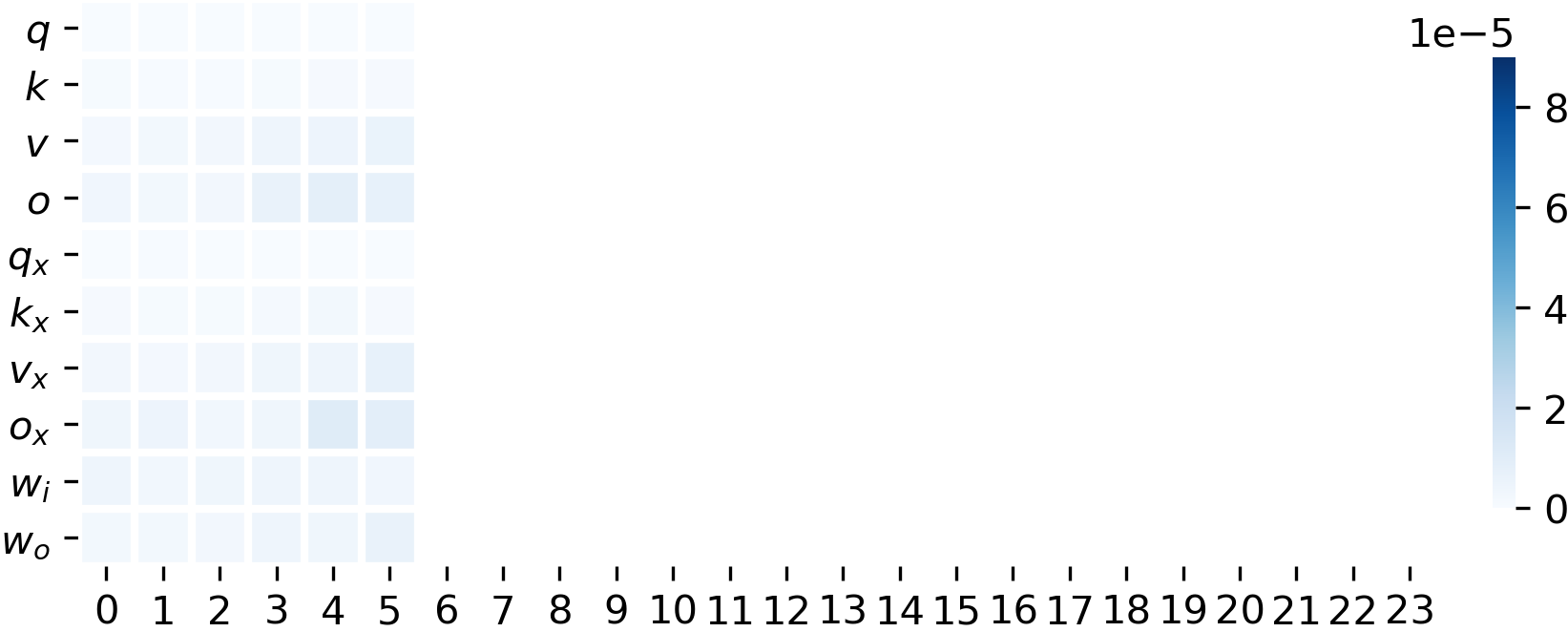}
  \caption{L1 change, Decoder, T5-Small, $n = 30$}
\end{figure}

\begin{figure}[t]
\centering
  \includegraphics[width=\linewidth]{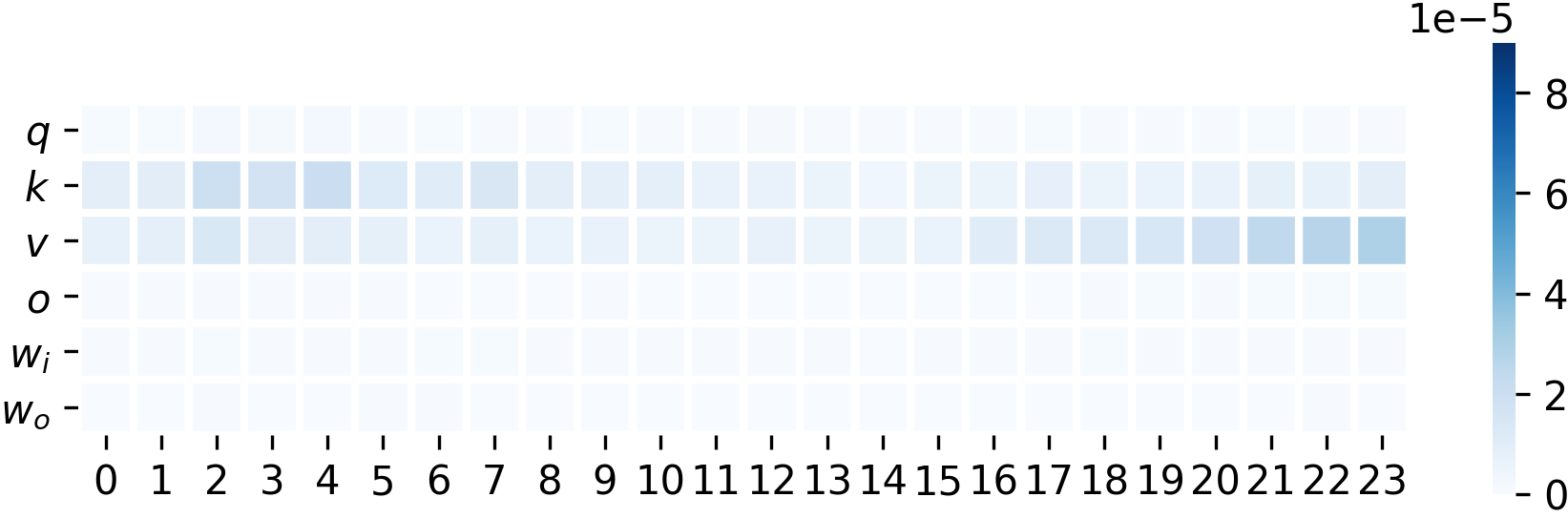}
  \caption{L1 change, Encoder, T5-11B, Shuffled Prompts, $n = 30$}
\end{figure}

\begin{figure}[t]
\centering
  \includegraphics[width=\linewidth]{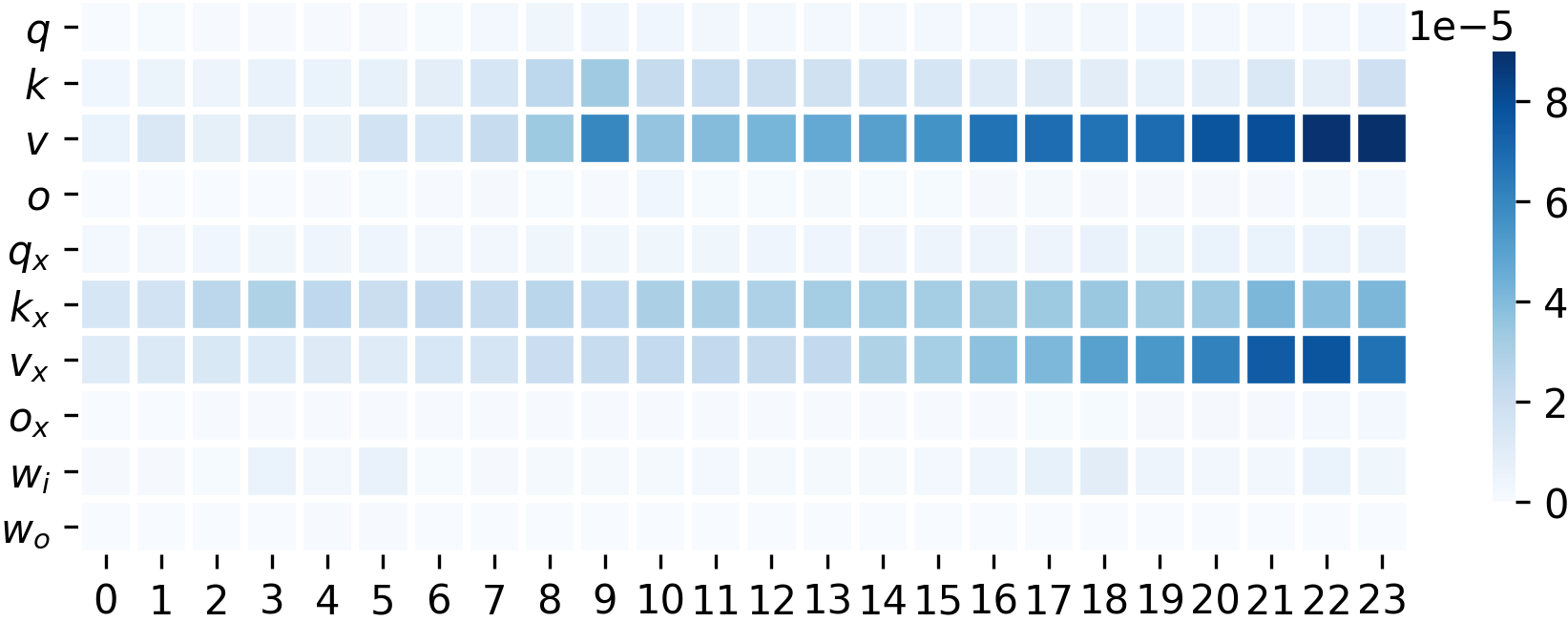}
  \caption{L1 change, Decoder, T5-11B, Shuffled Prompts, $n = 30$}
\end{figure}

\begin{figure}[t]
\centering
  \includegraphics[width=\linewidth]{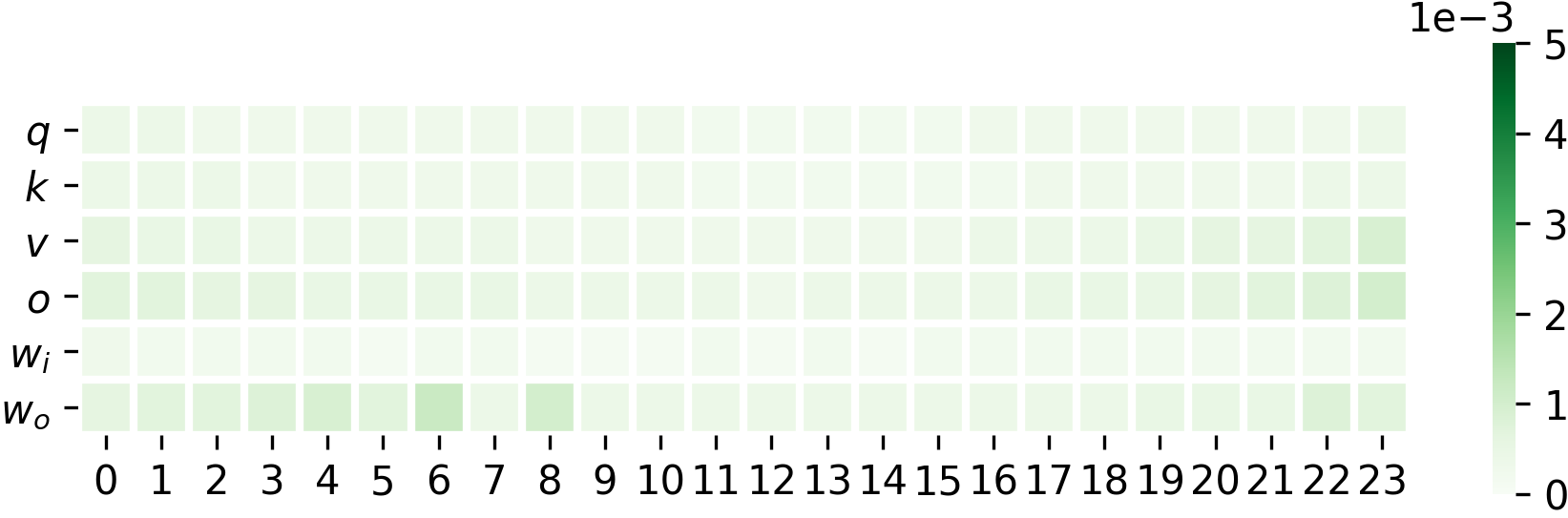}
  \caption{Angular change, Encoder, T5-11B, Shuffled Prompts, $n = 30$}
\end{figure}

